\documentclass[10pt,twocolumn,letterpaper]{article}

\usepackage[pagenumbers]{wacv}

\usepackage{url}            
\usepackage{booktabs}       
\usepackage{amsfonts}       
\usepackage{nicefrac}       
\usepackage{microtype}      
\usepackage{xcolor}         

\usepackage{graphicx}
\usepackage{amsmath}
\usepackage{amssymb}
\usepackage{array}
\usepackage{multirow}
\usepackage{color}
\usepackage{colortbl}
\usepackage{framed}
\usepackage{bm}
\usepackage{xspace}
\usepackage{enumitem}
\usepackage{xparse}
\usepackage{algorithm}
\usepackage{listings}
\usepackage{url}
\usepackage{mathtools}
\usepackage{pifont}
\usepackage{mathtools}
\usepackage{lipsum}
\usepackage{adjustbox}
\usepackage{bbm}
\usepackage{wrapfig}
\usepackage{algorithm}
\usepackage{algpseudocode}

\definecolor{darkred}{RGB}{212, 0, 0}
\definecolor{darkgreen}{RGB}{0, 128, 0}
\definecolor{Gray}{gray}{0.2}
\definecolor{lightgray}{gray}{0.92}
\definecolor{blond}{rgb}{0.98, 0.94, 0.75}
\definecolor{TitleColor}{gray}{0.95}
\definecolor{LightCyan}{rgb}{0.88,0.95,1}
\definecolor{OurColor}{rgb}{0.855, 0.937, 0.957}

\definecolor{blond}{rgb}{0.98, 0.94, 0.75}
\def \ie {\emph{i.e.}}
\def \eg {\emph{e.g.}}

\newcommand{\tit}[1]{\smallbreak\noindent\textbf{#1.}}
\newcommand{\tinytit}[1]{\noindent\textbf{#1.}}

\newcommand{\ours}{SPARK\xspace}

\newcommand{\inc}[1]{%
    \textcolor{darkgreen}{\textbf{\footnotesize$\Delta$+#1}}%
}

\newcommand{\dec}[1]{%
    \textcolor{darkgray!70}{\textbf{\footnotesize$\Delta$-#1}}%
}

\definecolor{wacvblue}{rgb}{0.21,0.49,0.74}
\usepackage[pagebackref,breaklinks,colorlinks,allcolors=wacvblue]{hyperref}

\def\confName{WACV}
\def\confYear{2027}

\title{SPARK: Input-Conditioned Sparse Activation Modulation for\\Frozen DiT-based Super-Resolution}

\author{
Federico Putamorsi$^{1,*}$,
Leonardo Zini$^{1,*}$,
Marcella Cornia$^{1}$,
Lorenzo Baraldi$^{1}$ \\
$^{1}$University of Modena and Reggio Emilia, Italy \\
$^{*}$Equal contribution. \\
{\tt\small\{name.surname\}@unimore.it}
\\\vspace{0.15cm}
}

\begin{document}
\maketitle

\begin{abstract}
Real-world image super-resolution (SR) increasingly relies on Diffusion Transformer (DiT) backbones, whose internal activations can be dominated by a small number of massive channels. Yet improving perceptual quality in these models still typically requires fine-tuning the network or attaching additional adapters, leaving this structured activation space largely unexplored for adaptation. We investigate whether dominant channels can instead serve as a compact adaptation interface for frozen DiT-based SR models. We first characterize their behavior in pretrained SR backbones and show through controlled interventions that they strongly affect reconstruction quality. Building on this observation, we introduce \textbf{\ours}, a lightweight input-conditioned controller that predicts bounded per-channel affine transformations for only the selected channels, while keeping the SR backbone and VAE frozen. Dominant channels are identified through an online activation-ranking procedure, and only a small predictor conditioned on the low-resolution VAE latent is optimized. Experiments on three DiT-based SR backbones across DIV2K, RealSR, and DRealSR show consistent gains in both fidelity and perceptual quality while modulating only eight channels per stream and block. Controlled comparisons further show that these gains cannot be explained by parameter budget or access to the selected channels alone.
\end{abstract}    

\begin{figure*}[t]
    \centering
    \includegraphics[width=.9\linewidth]{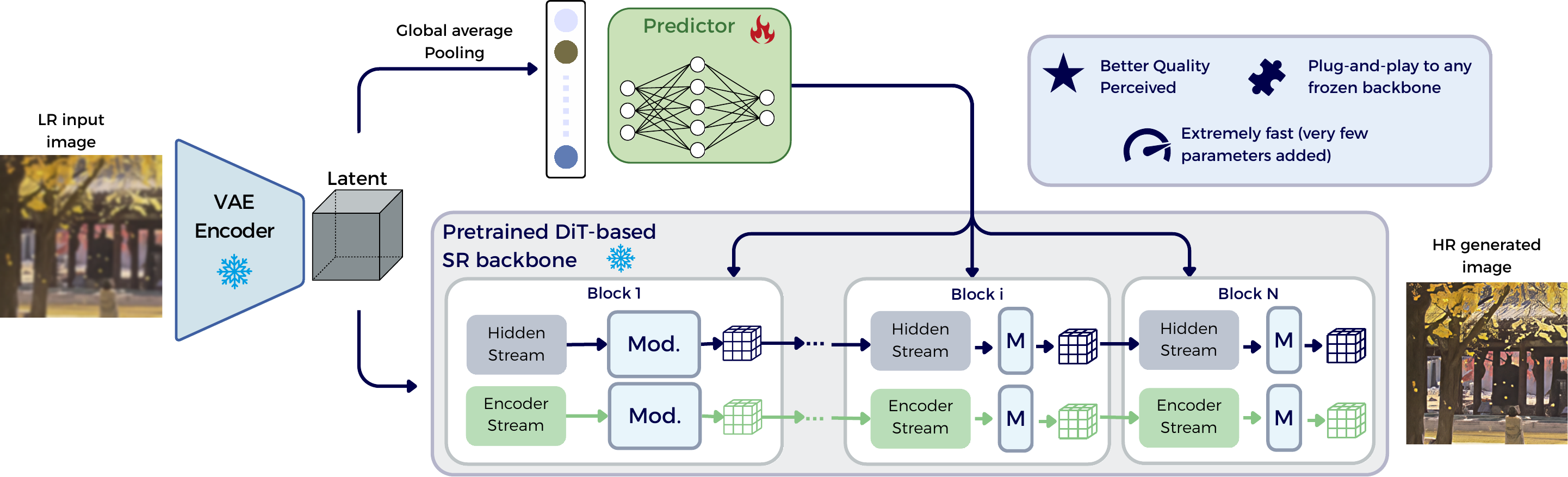}
    \caption{\textbf{Overview of the approach.} A frozen DiT-based SR backbone processes latent features extracted by a VAE encoder. A lightweight predictor, conditioned on a globally pooled latent representation, outputs per-channel affine modulation parameters that are injected into selected channels across multiple Transformer blocks. By modulating only a small subset of channels, the method improves perceptual quality without updating backbone weights, enabling a fast and plug-and-play enhancement strategy.}
    \label{fig:teaser}
    \vspace{-0.3cm}
\end{figure*}

\section{Introduction}
\label{sec:introduction}

Improving perceptual quality in single-image super-resolution (SR) remains challenging, particularly in real-world settings with complex and unknown degradations~\cite{wang2021real,wei2020component}. While early approaches focused on distortion-based objectives~\cite{dong2015image,lim2017enhanced}, modern methods increasingly rely on generative models~\cite{liu2022flow, esser2024scaling, ho2020denoising, lipman2024flowmatchingguidecode, Rombach_2022_CVPR} to synthesize realistic high-frequency details~\cite{ledig2017photo,wang2018esrgan}. Recent SR systems~\cite{sun2025pixel,lin2024diffbir,wu2024one} adopt Transformer-based generative backbones, including diffusion and flow-matching models~\cite{dong2025tsd, duan2025dit4sr, wu2025dp2osr, yi2025fine, sun2025pocketsr, cheng2025effective, fdoronzio2026gramsr}, achieving strong perceptual performance. However, how to effectively control perceptual quality in these models remains unclear. In particular, it is unknown whether such control requires modifying the full network, or whether it can be achieved by acting on a small and structured subset of the representation.

In this work, we address this question by analyzing internal activations of pretrained DiT-based SR models. We uncover a notable structural property: activation distributions are highly skewed, with a small subset of channels consistently reaching much larger magnitudes than the rest. Similar phenomena, referred to as \textit{massive activations}, have been observed in other Transformer-based models~\cite{sun2024massive, gan2025massive, darcet2024vision}. Through controlled ablation studies, we show that these dominant channels are not only statistically prominent but also functionally critical for reconstruction quality.

Motivated by this observation, we propose \textbf{\ours}, a lightweight activation-space modulation framework for DiT-based SR. \ours first identifies dominant channels in each block using an online ranking procedure that tracks an exponential moving average of channel importance and terminates as soon as the selected subset is stable, avoiding a full pass over the training data. It then learns an image-conditioned predictor that maps the low-resolution VAE latent to bounded affine parameters that modulate only the selected channels. The SR backbone and the VAE remain frozen, and only the predictor is optimized (Fig.~\ref{fig:teaser}). Our experiments show that modulating only eight channels per stream and block is sufficient to improve perceptual quality across multiple DiT-based SR backbones and benchmarks. Although trained on DIV2K~\cite{agustsson2017ntire}, the learned modulation generalizes to RealSR~\cite{cai2019toward} and DRealSR~\cite{wei2020component}. These results suggest that perceptual quality in DiT-based SR can be effectively controlled through a small set of dominant channels.

\tit{Contributions} To summarize, our contributions are as follows: \textit{(i)} We analyze massive activations in DiT-based SR models and show that dominant channels strongly affect reconstruction quality. \textit{(ii)} We introduce \ours, a frozen-backbone method that pairs an online, stability-terminated channel-selection procedure with an input-conditioned predictor of bounded affine parameters for the selected channels. \textit{(iii)} We demonstrate consistent improvements across three DiT-based SR backbones and multiple benchmarks, covering both fidelity and perceptual-quality metrics, and isolate the roles of channel selection, parameter budget, and modulation mechanism through controlled comparisons.

\section{Related Work}
\label{sec:related}

\tinytit{Early Super-Resolution Methods} 
Single-image super-resolution (SR) aims to recover a high-resolution (HR) image from a low-resolution (LR) input. Since SRCNN~\cite{dong2014learning}, deep learning has become the dominant paradigm, with early methods targeting synthetic degradations (\eg, bicubic downsampling) using pixel-wise reconstruction losses. Subsequent models improved performance through deeper architectures, residual learning, and multi-scale feature aggregation~\cite{ledig2017photo,lim2017enhanced}, but often generalized poorly to real-world noise, blur, and compression. This limitation motivated realistic degradation models and real-image benchmarks~\cite{cai2019toward,wei2020component}. In parallel, perceptual and adversarial approaches such as SRGAN~\cite{ledig2017photo} and ESRGAN~\cite{wang2018esrgan} shifted the focus toward perceptual quality. BSRGAN~\cite{zhang2021designing} and Real-ESRGAN~\cite{wang2021real} further improved realism through realistic degradation synthesis, although GAN-based SR remains susceptible to training instability and hallucinated details.

\tit{Diffusion-based Super-Resolution}
Diffusion models have emerged as a powerful generative paradigm across modalities~\cite{ho2020denoising,austin2021structured,bertolani2026diffusion}, and as a strong alternative for real-world SR due to their expressive generative priors and stable training. Methods such as StableSR~\cite{wang2024exploiting}, DiffBIR~\cite{lin2024diffbir}, PASD~\cite{yang2024pixel}, SeeSR~\cite{wu2024seesr}, and SUPIR~\cite{yu2024scaling} leverage large pre-trained diffusion priors to improve perceptual quality, texture realism, and semantic consistency.
Their multi-step denoising process, however, incurs substantial inference cost, motivating one- and few-step alternatives. OSEDiff~\cite{wu2024one} adopts latent score distillation for one-step SR, while TAD-SR~\cite{he2026one}, S3Diff~\cite{zhang2024degradation}, AddSR~\cite{tai2026addsr}, PiSA-SR~\cite{sun2025pixel}, and Gram-SR~\cite{fdoronzio2026gramsr} further improve efficiency and restoration quality through specialized supervision, degradation-aware conditioning, or adaptive refinement. These approaches generally adapt the generative prior or additional modules attached to it; in contrast, \ours keeps the SR backbone frozen and operates directly in its activation space.

\tit{Transformer and Flow-Matching Priors for SR} Recent SR methods increasingly adopt Transformer-based generative priors, offering strong global modeling and scalability. DiT-SR~\cite{cheng2025effective} introduces a hierarchical DiT model with frequency-aware conditioning, while DiT4SR~\cite{duan2025dit4sr} adapts large pre-trained priors by integrating LR guidance into attention and enhancing local detail recovery. TSD-SR~\cite{dong2025tsd} explores one-step restoration through distillation from a flow-matching text-to-image prior, highlighting the potential of modern generative models~\cite{liu2022flow,esser2024scaling,ho2020denoising,lipman2024flowmatchingguidecode,Rombach_2022_CVPR} for efficient SR. In a related direction, FluxSR~\cite{li2025one} builds on the FLUX~\cite{labs2025flux} Transformer prior and distills multi-step flow trajectories into a single-step model, reducing inference cost while preserving fidelity and perceptual quality. Rather than modifying these backbones, \ours keeps them frozen and improves SR through activation-space modulation.

\begin{figure*}[t]
\centering
\includegraphics[width=0.95\linewidth]{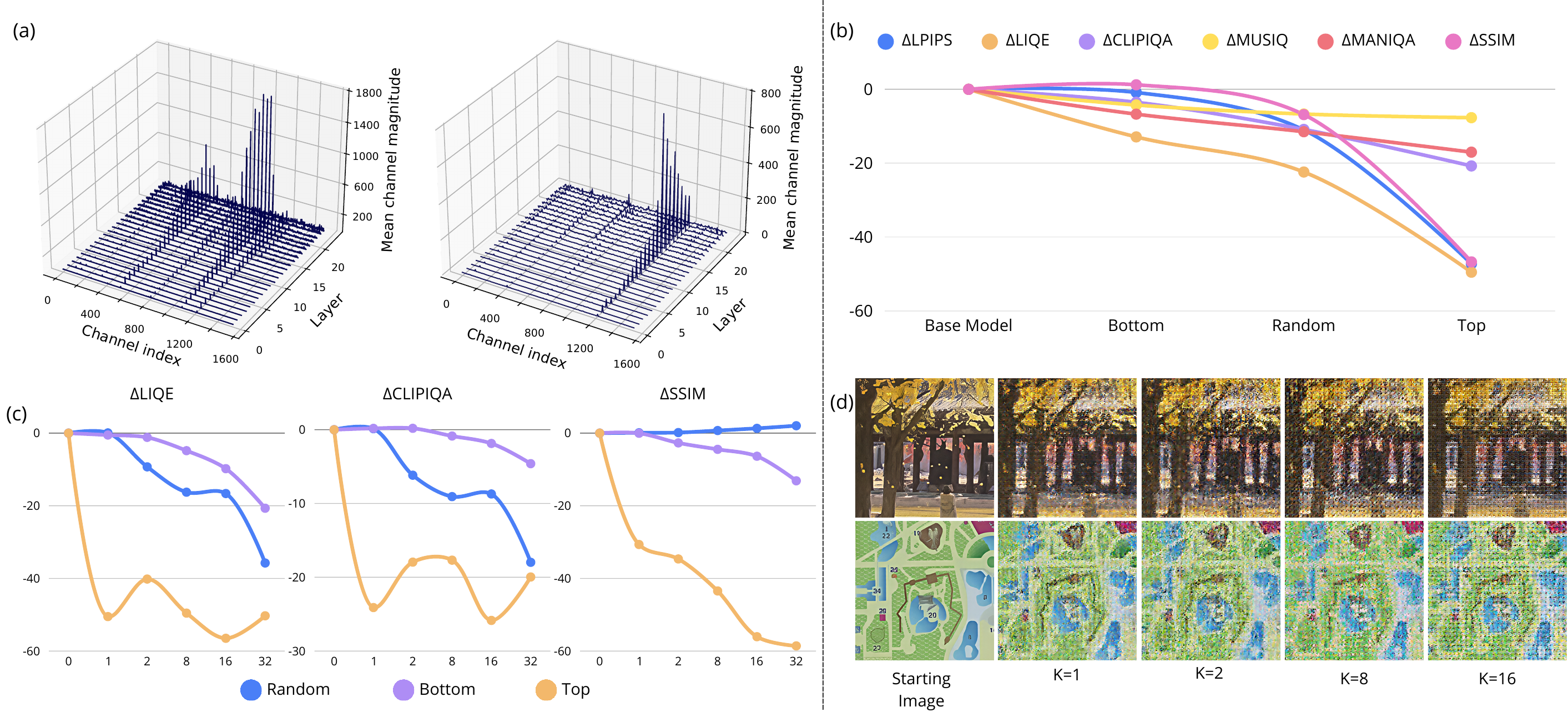}
\vspace{-0.3cm}
    \caption{\textbf{Analysis of activation concentration and its impact on reconstruction quality.}
(a) Per-channel activation magnitudes across blocks for DiT4SR~\cite{duan2025dit4sr} (left) and TSD-SR~\cite{dong2025tsd} (right), revealing a strongly skewed distribution with few dominant channels.
(b) Average performance drop when ablating top-, random-, and bottom-ranked channels (aggregated over $K$), highlighting the critical role of top-activation channels.
(c) Metric degradation as a function of $K$, showing strong sensitivity to top-$K$ removal and weak impact for random and bottom selections.
(d) Qualitative results for increasing $K$, where removing top channels progressively degrades perceptual quality.}
\label{fig:figure_analysis}
    \vspace{-0.3cm}
\end{figure*}

\tit{Massive Activations in DiTs} 
Massive activations in DiTs are abnormally large intermediate responses concentrated in a small set of channels, often dominating feature propagation and attention. Similar effects have been observed in LLMs and ViTs~\cite{sun2024massive, zhao2023unveiling}; in ViTs, they appear as attention artifacts in low-information regions and have been linked to positional embeddings~\cite{yang2024denoising,darcet2024vision}. Recent studies show that similar outliers also emerge in DiTs, causing unstable quantization and high distillation losses~\cite{fang2025tinyfusion,zhao2024vidit}. From a generative perspective, these dominant channels are strongly associated with fine-grained texture synthesis, while global semantic structure remains largely preserved under perturbations~\cite{gan2025massive}. Further analyses show that these channels encode spatially coherent information aligned with salient image regions, and that the subspace they span transfers across prompts~\cite{turri2026few}. This suggests a trade-off between texture fidelity and semantic consistency, and indicates that massive activations can be leveraged to extract semantically discriminative features for dense visual correspondence tasks~\cite{gan2025unleashing}.
These works establish the existence and functional relevance of massive activations, mainly through analysis or training-free manipulation. In contrast, we investigate whether the same sparse representation can serve as a \textit{learnable, input-conditioned adaptation interface} for frozen SR models. Our contribution lies in exploiting massive activations for representation-space adaptation, rather than in their discovery.
\section{Dominant Channels as an Adaptation Subspace}

Before introducing our adaptation mechanism, we examine whether magnitude-dominant channels form a compact set of effective intervention points in pretrained DiT-based SR models. We focus on two properties relevant to adaptation: activation concentration and intervention sensitivity.

\subsection{Preliminaries}
\label{sec:preliminaries}
We consider SR models built on Diffusion Transformers~\cite{peebles2023scalable}, and in particular on the multimodal variant~\cite{esser2024scaling} that operates on two parallel streams: a \textit{hidden} stream, which processes the input image as a sequence of patchified tokens, and an \textit{encoder} stream providing conditioning information. The two streams interact across Transformer blocks, allowing image content and conditioning signals to be jointly processed.

Let $\ell$ denote a DiT block, $s \in \{\mathrm{hid},\mathrm{enc}\}$ a stream, and $i$ an image index. Let $T_s$ and $C_s$ denote the number of tokens and channels in stream $s$, respectively. For image $i$, we denote the activation tensor at block $\ell$ and stream $s$ by $A_{i,s}^{(\ell)} \in \mathbb{R}^{T_s \times C_s}$.
We define the per-channel activation magnitude as the mean absolute activation over tokens,
\begin{equation}
    a_{i,s}^{(\ell)}(c)
    =
    \frac{1}{T_s}
    \sum_{t=1}^{T_s}
    \left|A_{i,s}^{(\ell)}(t,c)\right|,
    \label{eq:channel_mag}
\end{equation}
yielding
$a_{i,s}^{(\ell)} \in \mathbb{R}_{\geq 0}^{C_s}$,
which summarizes the activation magnitude of each channel.

We rank the entries of $a_{i,s}^{(\ell)}$ in decreasing order and define
\begin{equation}
    \mathrm{Top}_K\!\left(a_{i,s}^{(\ell)}\right)
    \subset \{1,\dots,C_s\}
\end{equation}
as the index set of the $K$ largest-magnitude channels. We refer to these as the \emph{dominant channels} of stream $s$ at block~$\ell$.

Finally, let $x$ denote the low-resolution input image, $z$ its latent encoding obtained from the frozen VAE encoder, $\hat{y}$ the reconstructed high-resolution output, and $y$ the corresponding target.

\subsection{Activation Analysis}
\label{sec:analysis}

\tinytit{Activation magnitude is highly concentrated across channels}
We first analyze the per-channel activation magnitudes
$a_{i,s}^{(\ell)}(c)$ (cf. Eq.~\eqref{eq:channel_mag}) across DiT blocks and streams. A clear and consistent pattern emerges: activation energy is highly unevenly distributed. A small subset of channels exhibits significantly larger activation magnitudes, while the majority remains barely active. Fig.~\ref{fig:figure_analysis} (a) illustrates this skewed activation landscape across blocks and channels.

To quantify this effect at the dataset level, we average the channel magnitudes over images for each block and stream, and measure the inequality of the resulting distribution using the Gini coefficient~\cite{10.1109/TIT.2009.2027527}. Focusing on the top-32 channels (~$2\%$ of the channel dimension), we observe high inequality (with a median Gini coefficient of $0.62$ across blocks), indicating a strong concentration of activation energy. Such a value reflects a highly skewed distribution, where a small subset of channels dominates the representation and likely plays a central role in determining the model output.

Although the ranking is computed from mean absolute activation magnitude, the resulting top-ranked channels also capture a disproportionate fraction of the squared activation energy. For $K=8$, they account on average for $85.6\%$ of~the encoder-stream energy and $68.8\%$ of the hidden-stream energy, while representing only $0.5\%$ of the channel dimension.

\tit{Dominant channels are critical and their effect scales with $K$}
Motivated by this observation, we next investigate the functional role of dominant channels through controlled analyses. For each block and stream, we zero subsets of channels selected according to their activation magnitude: the top-$K$ channels (highest magnitude), the bottom-$K$ channels (lowest magnitude), and randomly selected channels, while varying the number of channels $K$.  

We conduct this experiment using TSD-SR~\cite{dong2025tsd} on images from DRealSR~\cite{wei2020component}, and measure changes in SSIM~\cite{1284395}, LPIPS~\cite{zhang2018unreasonable}, LIQE~\cite{zhang2023blind}, MANIQA~\cite{yang2022maniqa}, MUSIQ~\cite{ke2021musiq}, and CLIP-IQA~\cite{wang2023exploring}. For each selection strategy, results are averaged across different values of $K$ to obtain an overall performance drop. As shown in Fig.~\ref{fig:figure_analysis} (b), removing the Top channels leads to a sharp degradation in reconstruction quality, whereas Random removal produces moderate degradation and Bottom removal has only a minor effect.

\tit{Metric-level sensitivity and qualitative effects}
We further analyze how reconstruction quality evolves as a function of the number of ablated channels $K$. As shown in Fig.~\ref{fig:figure_analysis} (c), removing top-$K$ channels causes a sharp, nearly monotonic degradation, with substantial drops even for small $K$ and no comparable trend for random or bottom-ranked channels. This behavior holds across perceptual and distortion-based metrics alike, indicating that the asymmetry is not metric-dependent but reflects a structural property of the model.

These effects are also visible qualitatively. As shown in Fig.~\ref{fig:figure_analysis} (d), progressively removing top-$K$ channels results in a gradual loss of perceptual quality as $K$ increases. Even small perturbations introduce noticeable artifacts, while larger values lead to severe degradation of fine details and structure. This further highlights the critical role of dominant channels and shows that perceptual quality is highly sensitive to a small number of high-activation features.

Overall, these results reveal a strong asymmetry in intervention sensitivity: magnitude-dominant channels provide substantially higher leverage over the reconstructed output than equally sized lower-ranked subsets. This motivates using them as a compact adaptation interface.
\section{Proposed Method}

\begin{figure*}[t]
    \centering
    \includegraphics[width=\linewidth]{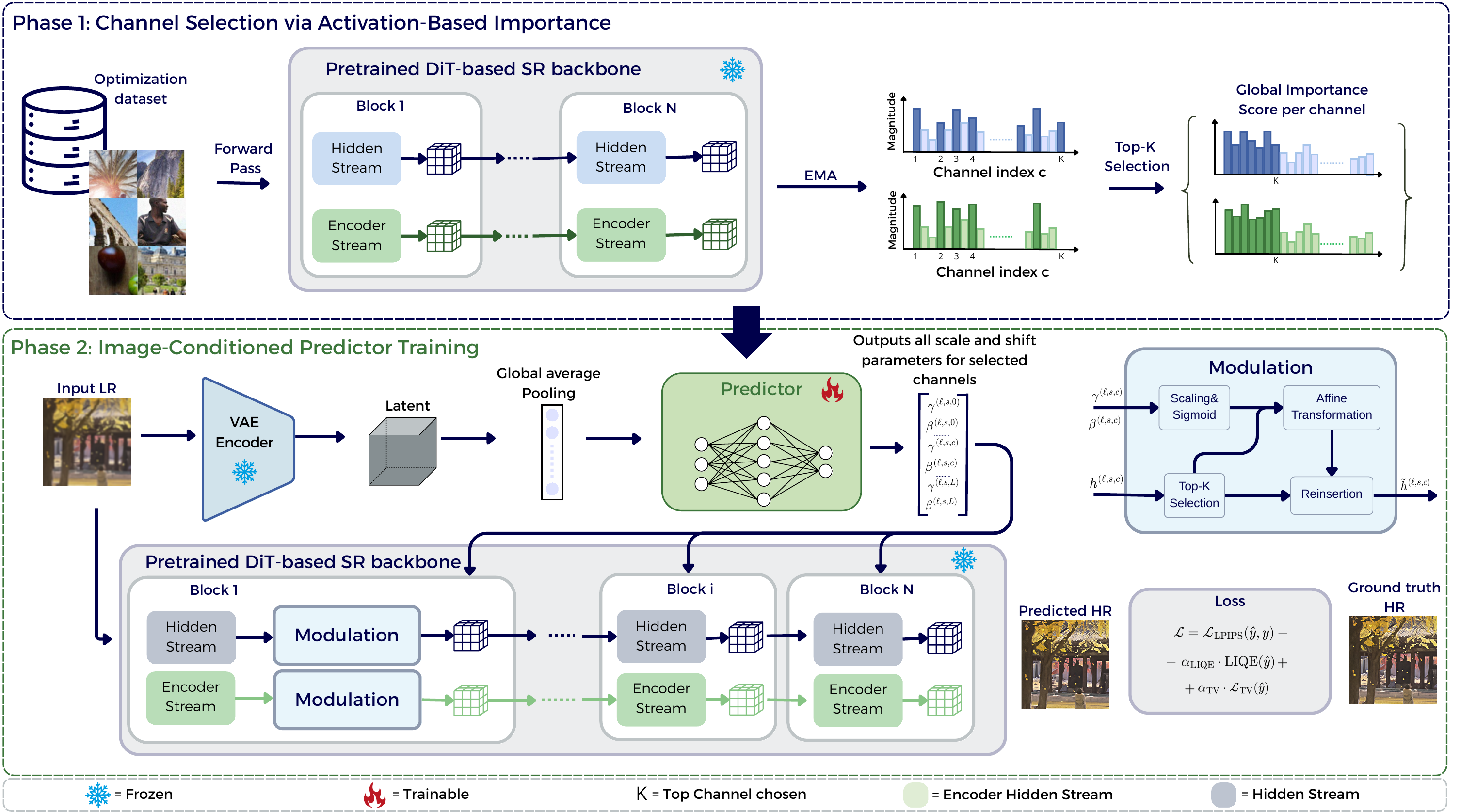}
    \vspace{-0.5cm}
    \caption{\textbf{Overview of the proposed activation-space modulation framework.} Phase 1 (top) identifies dominant channels by aggregating activation magnitudes over a dataset and selecting the top-$K$ channels per block and stream. Phase 2 (bottom) trains a lightweight predictor that, given a VAE latent, outputs scale and shift parameters applied to the selected channels of a frozen DiT-based SR backbone. The modulation is injected into both hidden and encoder streams via feature-wise affine transformations. Only the predictor is trained, enabling efficient, input-conditioned control of reconstruction quality.}
    \label{fig:method}
    \vspace{-0.5cm}
\end{figure*}

We propose \ours, a modulation approach for DiT-based SR that improves reconstruction quality through selective feature adaptation. Our analysis (cf. Sec.~\ref{sec:analysis}) shows that a small subset of channels concentrates most of the activation energy and strongly influences the output. We therefore intervene only in this sparse subspace, learning modulation parameters for the selected channels. To control their contribution, we apply feature-wise affine transformations, enabling efficient amplification or attenuation of dominant channels without modifying the pretrained backbone.

Our approach follows two phases: \textit{(i)} channel selection via activation-based importance, where we identify the most influential channels in each Transformer block using dataset-level statistics, and \textit{(ii)} image-conditioned predictor training, where a lightweight module maps input features to modulation parameters applied to the selected channels. At inference time, the predictor generates conditioned parameters from the VAE latent, enabling efficient enhancement without iterative optimization.

\subsection{Phase 1: Channel Selection via Activation-based Importance}
We aim to identify a small subset of channels that consistently exhibit high activation magnitude across inputs, yielding a compact control space for modulation. Rather than computing channel statistics over the full optimization dataset, we estimate channel importance in an online and sequential manner. Given a pretrained SR backbone, we process optimization samples in mini-batches. For each block $\ell$ and each stream $s$ described in Sec.~\ref{sec:preliminaries}, we compute the per-channel activation magnitude $a_t^{(\ell,s)}$, averaged over the mini-batch at iteration $t$ following Eq.~\eqref{eq:channel_mag}, and maintain an exponential moving average (EMA) of channel importance:
\begin{equation}
m_{t}^{(\ell,s)} = \lambda\, m_{t-1}^{(\ell,s)} + (1-\lambda)\,a_{t}^{(\ell,s)},
\end{equation}
where $\lambda \in [0,1)$ controls the memory of the estimator. This online update smooths batch-level fluctuations while progressively refining the channel ranking. At each iteration, we rank channels according to their current EMA score $m_{t}^{(\ell,s)}$ and define the provisional top-$K$ set
\begin{equation}
\mathcal{I}_{t}^{(\ell,s)} \subset \{1,\dots,C_s\}.
\end{equation}

Instead of selecting channels after a fixed full pass over the dataset, we monitor the stability of the ranking across snapshots taken every $W$ update steps (with $W$ equal to $5$ in our experiments). Let $\mathcal{I}_{w}^{(\ell,s)}$ denote the provisional top-$K$ set obtained from the EMA scores $m_{wW}^{(\ell,s)}$ at the end of the $w$-th window. We stop the estimation procedure once the selected subset remains unchanged, or nearly unchanged, across $P$ consecutive window transitions:
\begin{equation}
\frac{1}{P}\sum_{p=0}^{P-1}
\frac{\left| \mathcal{I}_{w-p}^{(\ell,s)} \cap \mathcal{I}_{w-p-1}^{(\ell,s)} \right|}
{\left| \mathcal{I}_{w-p}^{(\ell,s)} \cup \mathcal{I}_{w-p-1}^{(\ell,s)} \right|}
\geq \tau,
\label{eq:stability}
\end{equation}
where $\tau$ is a stability threshold, set to $0.9$ in all experiments. We use $\lambda=0.95$ for the EMA and $P=4$ consecutive windows. This sequential screening procedure yields a compact set of informative channels without requiring an exhaustive dataset-level pass. Importantly, it adapts naturally to the empirical convergence of channel statistics: once the importance ranking stabilizes, further samples provide diminishing returns and the procedure terminates early. After convergence, we retain the final top-$K$ channels for each block and stream:
\begin{equation}
\mathcal{S} = \{(\ell, s, \mathcal{I}^{(\ell,s)})\},
\end{equation}
where $K=8$ channels are selected per stream and block in our implementation. This sparse channel map defines the modulation subspace for Phase 2.

\subsection{Phase 2: Image-Conditioned Predictor}
Given the selected channel subset $\mathcal{S}$, we learn an image-conditioned predictor that outputs modulation parameters for these channels. This formulation decouples channel selection (where to intervene) from modulation (how to intervene). For each selected channel $c \in \mathcal{I}^{(\ell,s)}$, we apply a feature-wise affine modulation:
\begin{equation}
\tilde{h}^{(\ell,s,c)}(x) = \gamma^{(\ell,s,c)}(x) \cdot h^{(\ell,s,c)}(x) + \beta^{(\ell,s,c)}(x),
\end{equation}
where $h^{(\ell,s,c)}$ denotes the activation of channel $c$ at block $\ell$ and stream $s$, and $\gamma^{(\ell,s,c)}$, $\beta^{(\ell,s,c)}$ are input-dependent scale and shift parameters predicted by the modulation network.

To predict these parameters, we encode the low-resolution input image $x$ using the same frozen VAE~\cite{Rombach_2022_CVPR} employed by the SR backbone, producing a latent representation $z$. Since this encoding is already available in the SR pipeline, conditioning reuses the existing VAE latent and introduces only the overhead of the lightweight predictor. We then compute a compact global feature vector via spatial pooling (channel-wise mean), which provides sufficient information to guide image-dependent modulation while keeping the predictor lightweight. The feature vector is passed to an MLP $f_{\theta}$, which outputs all scale and shift parameters:
\begin{equation}
\bigl(
\bar{\gamma}^{(\ell,s,c)}(x),
\bar{\beta}^{(\ell,s,c)}(x)
\bigr)_{\ell,s,\,c\in\mathcal{I}^{(\ell,s)}}
=
f_{\theta}\!\left(\operatorname{AvgPool}(z)\right).
\end{equation}
 
The output dimensionality of the MLP is determined by the total number of selected channels across all blocks and streams:
\begin{equation}
\mathrm{dim}_{\text{out}} = 2 \sum_{\ell,s} |\mathcal{I}^{(\ell,s)}|,
\end{equation}
where the factor of 2 reflects that, for each selected channel $c \in \mathcal{I}^{(\ell,s)}$, the predictor outputs an individual scale--shift pair $\bigl(\bar{\gamma}^{(\ell,s,c)},\bar{\beta}^{(\ell,s,c)}\bigr)$. Thus, modulation parameters are predicted independently for every selected channel in every stream of every block.

To ensure stable modulation, we constrain the predicted parameters to lie within fixed intervals. 
Specifically, we apply a sigmoid followed by linear scaling:
\begin{equation}
\begin{aligned}
\gamma^{(\ell,s,c)}(x)
&=
\gamma_{\min}
+
(\gamma_{\max}-\gamma_{\min})
\cdot
\sigma\!\left(\bar{\gamma}^{(\ell,s,c)}(x)\right),
\\
\beta^{(\ell,s,c)}(x)
&=
\beta_{\min}
+
(\beta_{\max}-\beta_{\min})
\cdot
\sigma\!\left(\bar{\beta}^{(\ell,s,c)}(x)\right),
\end{aligned}
\end{equation}

where $\bar{\gamma}$ and $\bar{\beta}$ denote the unconstrained outputs of $f_\theta$, and $\sigma(\cdot)$ is the sigmoid function. Thanks to the sparse selection, the output dimensionality remains small compared with full-channel modulation, keeping the predictor lightweight.

During training, both the SR backbone and the VAE are frozen; only the predictor parameters $\theta$ are optimized. The predicted parameters are injected into the forward pass, allowing gradients to propagate through the SR model back to the predictor. The training objective combines perceptual fidelity via LPIPS~\cite{zhang2018unreasonable}, perceptual quality via LIQE~\cite{zhang2023blind}, and total variation regularization:
\begin{equation}
\mathcal{L} =
 \mathcal{L}_{\mathrm{LPIPS}}(\hat{y}, y)
\;-\;
\alpha_{\mathrm{LIQE}} \cdot \mathrm{LIQE}(\hat{y})
\;+\;
\alpha_{\mathrm{TV}} \cdot \mathcal{L}_{\mathrm{TV}}(\hat{y}),
\end{equation}
where $\hat{y}$ is the reconstructed high-resolution image. We set $\alpha_{\mathrm{LIQE}} = 0.1$ and $\alpha_{\mathrm{TV}} = 0.0001$ to balance loss scales and ensure stable training.

Overall, this phase learns a compact, input-conditioned control mechanism over a sparse subset of channels. Despite modulating only $K$ channels per stream and block, the resulting controller consistently improves perceptual quality over the frozen baseline, showing that a small set of dominant features is sufficient to steer SR generation. This provides an efficient adaptation mechanism that operates on only a small fraction of the representation while leaving the pretrained SR backbone unchanged.
\section{Experiments}
\label{sec:experiments}

\subsection{Experimental Setup}

\tinytit{Model variants and baselines}
We evaluate our approach on three DiT-based SR models: TSD-SR~\cite{dong2025tsd}, DiT4SR~\cite{duan2025dit4sr}, and TEASR~\cite{gao2026teasr}. These models provide complementary testbeds for evaluating whether sparse activation adaptation transfers across independently developed SR architectures. For each model, we use the publicly released implementation and checkpoint, and take the corresponding frozen, unmodified backbone as the baseline.

\tit{Implementation details}
We instantiate $f_{\theta}$ as a lightweight MLP with two hidden layers of width 256 and SiLU activations, mapping the pooled VAE latent to per-channel affine parameters. We constrain the modulation ranges to $\gamma \in [0.5,1.5]$ and $\beta \in [-0.2,0.2]$ in all experiments. The predictor is trained on the DIV2K training set~\cite{agustsson2017ntire}, using synthetic degradations generated with the Real-ESRGAN pipeline~\cite{wang2021real}. We train for one epoch with batch size 16 using the Adam optimizer~\cite{kingma2014adam} and a learning rate of $1\times10^{-4}$, with early stopping based on validation performance. No additional data augmentation or external data is used. All experiments are conducted on NVIDIA L40S GPUs with 48 GB of memory. The SR backbone with the VAE encoder and decoder remain frozen, and only the predictor is optimized, isolating the effect of activation-space modulation.

\tit{Evaluation protocol}
We evaluate on DIV2K validation~\cite{agustsson2017ntire}, DRealSR~\cite{wei2020component}, and RealSR~\cite{cai2019toward}, covering both synthetic and real-world degradations. For the real-world datasets, each LR image is center-cropped to $128\times128$, and the corresponding aligned HR region is used for evaluation. This protocol evaluates both in-domain performance on DIV2K validation and generalization from synthetic training degradations to the real-world distributions of RealSR and DRealSR.

We report fidelity metrics, SSIM~\cite{1284395} and LPIPS~\cite{zhang2018unreasonable}, together with no-reference perceptual-quality metrics, including MANIQA~\cite{yang2022maniqa}, MUSIQ~\cite{ke2021musiq}, CLIP-IQA~\cite{wang2023exploring}, TOPIQ~\cite{topiq}, and LIQE~\cite{zhang2023blind}\footnote{For readability, SSIM, LPIPS, MANIQA, CLIP-IQA, and TOPIQ are reported $\times100$ in all tables.}.

\begin{table*}[t]
\centering
\footnotesize
\setlength{\tabcolsep}{1.15pt}

\begin{tabular}{
@{}
c                 
p{0.85em}          
l                 
p{1.2em}          
r l p{0.85em}      
r l p{1.2em}     
r l p{0.85em}    
r l p{0.85em}     
r l p{0.85em}      
r l p{0.85em}      
>{\columncolor{lightgray!50}}r   
>{\columncolor{lightgray!50}}l   
@{}
}

\toprule

&
&
&
&
\multicolumn{5}{c}{\textbf{Fidelity}}
&
&
\multicolumn{14}{c}{\textbf{Perceptual Quality}}
\\

\cmidrule(lr){5-9}
\cmidrule(lr){11-24}

&
&
&
&
\multicolumn{2}{c}{SSIM $\uparrow$}
&
&
\multicolumn{2}{c}{LPIPS $\downarrow$}
&
&
\multicolumn{2}{c}{MANIQA $\uparrow$}
&
&
\multicolumn{2}{c}{MUSIQ $\uparrow$}
&
&
\multicolumn{2}{c}{CLIP-IQA $\uparrow$}
&
&
\multicolumn{2}{c}{TOPIQ $\uparrow$}
&
&
\multicolumn{2}{c}{\cellcolor{lightgray!50}LIQE $\uparrow$}
\\

\midrule


&
&
TSD-SR~\cite{dong2025tsd}
&
&
71.68 & &
&
31.11 & &
&
58.12 & &
&
66.01 & &
&
73.63 & &
&
62.49 & &
&
4.05 &
\\

\rowcolor{OurColor}
\cellcolor{white}
&
\cellcolor{white}
&
\textbf{\quad$+$ \ours (Ours)}
&
&
\textbf{73.94} & \inc{2.26}
&
&
\textbf{31.07} & \inc{0.04}
&
&
\textbf{60.13} & \inc{2.01}
&
&
\textbf{68.16} & \inc{2.15}
&
&
\textbf{76.32} & \inc{2.69}
&
&
\textbf{67.39} & \inc{4.90}
&
&
\cellcolor{lightgray!50}\textbf{4.47}
&
\cellcolor{lightgray!50}\inc{0.42}
\\

\cmidrule(lr){3-24}

&
&
DiT4SR~\cite{duan2025dit4sr}
&
&
61.02 & &
&
43.72 & &
&
60.88 & &
&
65.20 & &
&
69.36 & &
&
58.50 & &
&
4.08 &
\\

\rowcolor{OurColor}
\cellcolor{white}
&
\cellcolor{white}
&
\textbf{\quad$+$ \ours (Ours)}
&
&
\textbf{66.26} & \inc{5.24}
&
&
\textbf{37.18} & \inc{6.54}
&
&
\textbf{62.99} & \inc{2.11}
&
&
\textbf{66.06} & \inc{0.86}
&
&
\textbf{71.38} & \inc{2.02}
&
&
\textbf{58.67} & \inc{0.17}
&
&
\cellcolor{lightgray!50}\textbf{4.20}
&
\cellcolor{lightgray!50}\inc{0.12}
\\

\cmidrule(lr){3-24}

&
&
TEASR~\cite{gao2026teasr}
&
&
73.07 & &
&
30.97 & &
&
56.17 & &
&
63.10 & &
&
56.90 & &
&
57.42 & &
&
3.09 &
\\

\rowcolor{OurColor}
\cellcolor{white}
\multirow{-7}{*}{%
    \rotatebox[origin=c]{90}{\textbf{DRealSR}}%
}
&
\cellcolor{white}
&
\textbf{\quad$+$ \ours (Ours)}
&
&
\textbf{75.71} & \inc{2.64}
&
&
\textbf{30.84} & \inc{0.13}
&
&
\textbf{61.42} & \inc{5.25}
&
&
\textbf{67.73} & \inc{4.63}
&
&
\textbf{65.39} & \inc{8.49}
&
&
\textbf{65.14} & \inc{7.72}
&
&
\cellcolor{lightgray!50}\textbf{4.10}
&
\cellcolor{lightgray!50}\inc{1.01}
\\

\midrule


&
&
TSD-SR~\cite{dong2025tsd}
&
&
68.84 & &
&
\textbf{27.96} & &
&
63.03 & &
&
70.72 & &
&
72.55 & &
&
66.40 & &
&
4.19 &
\\

\rowcolor{OurColor}
\cellcolor{white}
&
\cellcolor{white}
&
\textbf{\quad$+$ \ours (Ours)}
&
&
\textbf{70.43} & \inc{1.59}
&
&
28.44 & \dec{0.48}
&
&
\textbf{64.64} & \inc{1.61}
&
&
\textbf{72.14} & \inc{1.42}
&
&
\textbf{75.35} & \inc{2.80}
&
&
\textbf{69.83} & \inc{3.43}
&
&
\cellcolor{lightgray!50}\textbf{4.69}
&
\cellcolor{lightgray!50}\inc{0.50}
\\

\cmidrule(lr){3-24}

&
&
DiT4SR~\cite{duan2025dit4sr}
&
&
66.05 & &
&
33.35 & &
&
59.56 & &
&
63.63 & &
&
62.24 & &
&
54.81 & &
&
3.63 &
\\

\rowcolor{OurColor}
\cellcolor{white}
&
\cellcolor{white}
&
\textbf{\quad$+$ \ours (Ours)}
&
&
\textbf{69.27} & \inc{3.22}
&
&
\textbf{28.51} & \inc{4.84}
&
&
\textbf{61.70} & \inc{2.14}
&
&
\textbf{64.78} & \inc{1.15}
&
&
\textbf{64.66} & \inc{2.42}
&
&
\textbf{55.19} & \inc{0.38}
&
&
\cellcolor{lightgray!50}\textbf{3.74}
&
\cellcolor{lightgray!50}\inc{0.11}
\\

\cmidrule(lr){3-24}

&
&
TEASR~\cite{gao2026teasr}
&
&
69.62 & &
&
\textbf{27.28} & &
&
58.98 & &
&
67.31 & &
&
54.83 & &
&
60.29 & &
&
3.28 &
\\

\rowcolor{OurColor}
\cellcolor{white}
\multirow{-7}{*}{%
    \rotatebox[origin=c]{90}{\textbf{RealSR}}%
}
&
\cellcolor{white}
&
\textbf{\quad$+$ \ours (Ours)}
&
&
\textbf{71.59} & \inc{1.97}
&
&
28.61 & \dec{1.33}
&
&
\textbf{64.68} & \inc{5.70}
&
&
\textbf{71.05} & \inc{3.74}
&
&
\textbf{62.27} & \inc{7.44}
&
&
\textbf{68.08} & \inc{7.79}
&
&
\cellcolor{lightgray!50}\textbf{4.17}
&
\cellcolor{lightgray!50}\inc{0.89}
\\

\midrule


&
&
TSD-SR~\cite{dong2025tsd}
&
&
55.95 & &
&
27.35 & &
&
60.63 & &
&
70.58 & &
&
71.83 & &
&
66.60 & &
&
4.27 &
\\

\rowcolor{OurColor}
\cellcolor{white}
&
\cellcolor{white}
&
\textbf{\quad$+$ \ours (Ours)}
&
&
\textbf{58.72} & \inc{2.77}
&
&
\textbf{26.78} & \inc{0.57}
&
&
\textbf{62.45} & \inc{1.82}
&
&
\textbf{71.99} & \inc{1.41}
&
&
\textbf{75.77} & \inc{3.94}
&
&
\textbf{70.36} & \inc{3.76}
&
&
\cellcolor{lightgray!50}\textbf{4.62}
&
\cellcolor{lightgray!50}\inc{0.35}
\\

\cmidrule(lr){3-24}

&
&
DiT4SR~\cite{duan2025dit4sr}
&
&
54.10 & &
&
34.48 & &
&
59.51 & &
&
66.42 & &
&
67.59 & &
&
57.71 & &
&
4.03 &
\\

\rowcolor{OurColor}
\cellcolor{white}
&
\cellcolor{white}
&
\textbf{\quad$+$ \ours (Ours)}
&
&
\textbf{55.27} & \inc{1.17}
&
&
\textbf{31.23} & \inc{3.25}
&
&
\textbf{62.27} & \inc{2.76}
&
&
\textbf{66.87} & \inc{0.45}
&
&
\textbf{71.55} & \inc{3.96}
&
&
\textbf{60.10} & \inc{2.39}
&
&
\cellcolor{lightgray!50}\textbf{4.25}
&
\cellcolor{lightgray!50}\inc{0.22}
\\

\cmidrule(lr){3-24}

&
&
TEASR~\cite{gao2026teasr}
&
&
56.72 & &
&
29.41 & &
&
57.42 & &
&
67.37 & &
&
54.72 & &
&
57.06 & &
&
3.52 &
\\

\rowcolor{OurColor}
\cellcolor{white}
\multirow{-7}{*}{%
    \rotatebox[origin=c]{90}{\textbf{DIV2K Val}}%
}
&
\cellcolor{white}
&
\textbf{\quad$+$ \ours (Ours)}
&
&
\textbf{59.31} & \inc{2.59}
&
&
\textbf{28.66} & \inc{0.75}
&
&
\textbf{63.16} & \inc{5.74}
&
&
\textbf{71.51} & \inc{4.14}
&
&
\textbf{64.94} & \inc{10.22}
&
&
\textbf{66.90} & \inc{9.84}
&
&
\cellcolor{lightgray!50}\textbf{4.37}
&
\cellcolor{lightgray!50}\inc{0.85}
\\

\bottomrule
\end{tabular}

\vspace{-0.15cm}

\caption{
\textbf{Comparison of DiT-based SR backbones across different datasets.} For each metric, the score is followed by its direction-normalized change
with respect to the corresponding frozen backbone
($\Delta+$: improvement; $\Delta-$: degradation).
LIQE is shaded in gray because it is explicitly included in the training objective of \ours.
}

\label{tab:comparison}
\vspace{-0.45cm}

\end{table*}

\subsection{Experimental Results}

\begin{figure}[t]
\vspace{-0.1cm}
    \centering
    \includegraphics[width=0.98\linewidth]{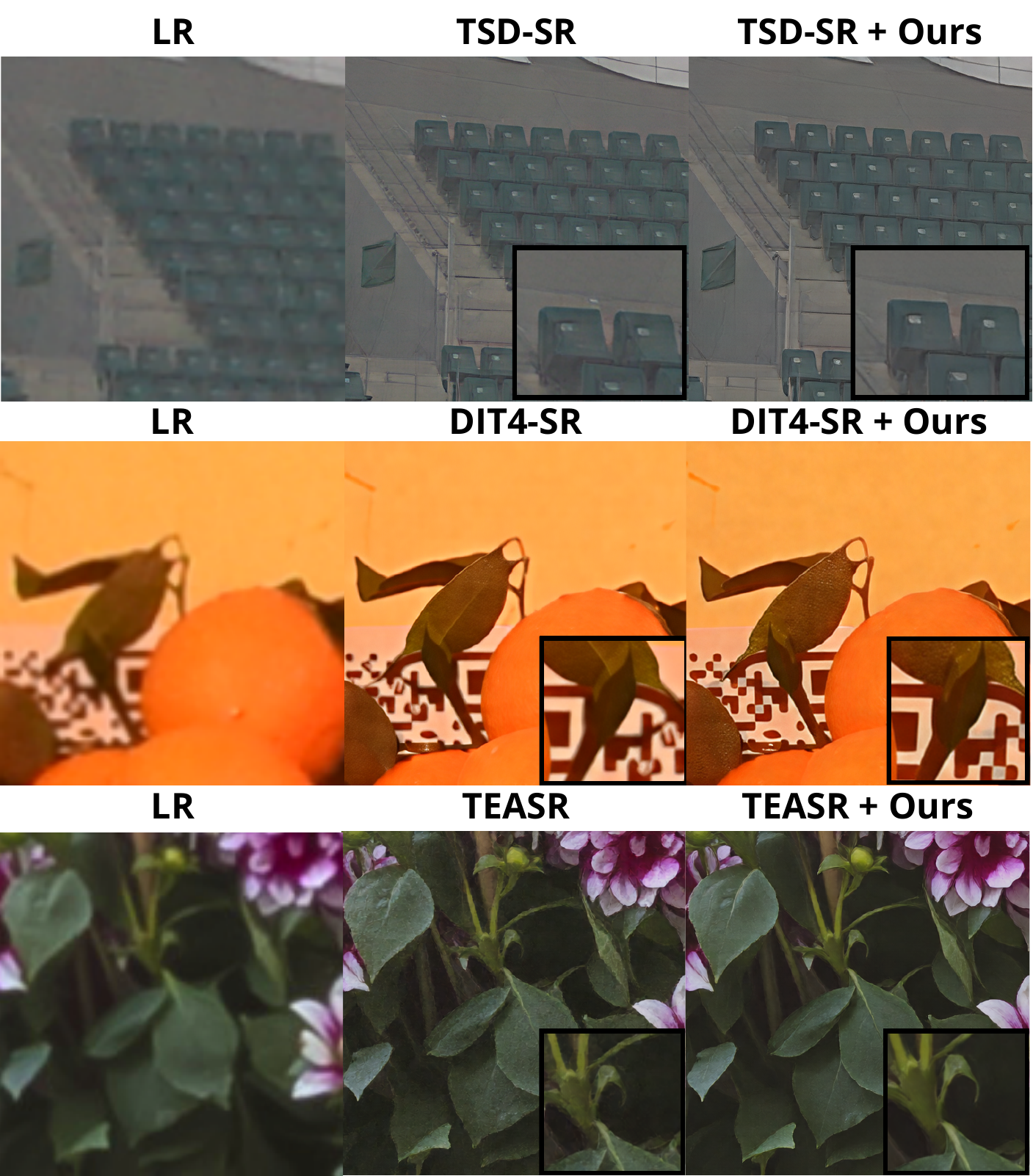}
    \vspace{-0.2cm}
    \caption{\textbf{Qualitative comparison across DiT-based SR backbones.} From left to right: LR input, frozen baseline, and \ours.}
    \label{fig:qualitative_competitors}
    \vspace{-0.55cm}
\end{figure}

\tit{Main results} 
Table~\ref{tab:comparison} reports results on DRealSR, RealSR, and DIV2K-Val across the three considered DiT-based SR backbones. \ours improves 61 of 63 metric-dataset-backbone combinations, with gains extending beyond no-reference perceptual quality: SSIM improves in all nine settings and LPIPS in seven. Despite being trained only on synthetically degraded images, the learned modulation transfers consistently to both real-world benchmarks without adapting the SR backbones. The trend is particularly strong for TEASR, where \ours improves 20 of 21 scores, with gains up to $+5.74$ MANIQA, $+4.63$ MUSIQ, $+10.22$ CLIP-IQA, and $+9.84$ TOPIQ. Qualitative results in Fig.~\ref{fig:qualitative_competitors} further show sharper local structures and details across all three backbones. Overall, these results show that a small activation-space intervention can improve diverse DiT-based SR models while largely preserving or improving fidelity.

\tit{Comparison with alternative adaptation strategies}
To disentangle the effects of channel selection, parameter budget, and modulation mechanism, in Table~\ref{tab:intervention_ablation} we compare \ours with five controlled parameter-efficient baselines on DRealSR. All methods are trained using the same data, objective, optimization, and frozen SR backbone (\ie, TSD-SR). Specifically, IA$^3$~\cite{liu2022few}, Houlsby adapters~\cite{houlsby2019parameter}, LoReFT~\cite{wu2024reft}, and channel-localized LoRA~\cite{hu2022lora} are restricted to the same top-$8$ channels per stream and block selected in Phase~1, providing alternative adaptation mechanisms over the same activation subspace. We additionally include a parameter-matched LoRA baseline, which applies standard LoRA to the feed-forward projections using a comparable trainable parameter budget. Together, these controls test whether the gains arise simply from intervening on the dominant channels or from using a similar adaptation capacity.

The results show that the dominant-channel subspace provides a consistently effective adaptation interface: nearly all alternative mechanisms improve over the frozen baseline when restricted to the same selected channels. Among these shared-subspace methods, \ours achieves the strongest MUSIQ, CLIP-IQA, TOPIQ, and LIQE scores, while LoReFT obtains the best MANIQA. Parameter-matched LoRA instead favors fidelity, achieving the best SSIM and LPIPS, but remains below \ours on MUSIQ, CLIP-IQA, and TOPIQ. Overall, these results support two complementary conclusions: the selected dominant channels are broadly useful for adaptation, and the gains of \ours cannot be explained by a comparable parameter budget alone.

\tit{Importance of channel selection}
Table~\ref{tab:selection_ablation} compares different channel selection strategies under the same modulation budget ($K=8$). Modulating the top-activation channels yields the strongest perceptual quality, achieving the best MANIQA, MUSIQ, CLIP-IQA, TOPIQ, and LIQE scores, while also providing the best SSIM. Random selection achieves the best LPIPS, indicating a slightly different fidelity--perception trade-off. Overall, dominant channels provide the most effective subspace for our objective of improving perceptual SR quality.

\begin{table}[t]
\centering
\footnotesize
\setlength{\tabcolsep}{0.5pt}
\resizebox{\linewidth}{!}{
\begin{tabular}{
@{}
l
cc
ccccc
>{\columncolor{lightgray!50}}c
@{}
}

\toprule

&
\multicolumn{2}{c}{\textbf{Fidelity}}
&
\multicolumn{5}{c}{\textbf{Perceptual Quality}}
\\

\cmidrule(lr){2-3}
\cmidrule(lr){4-8}

&
{\scriptsize SSIM $\uparrow$}
&
{\scriptsize LPIPS $\downarrow$}
&
{\scriptsize MANIQA $\uparrow$}
&
{\scriptsize MUSIQ $\uparrow$}
&
{\scriptsize CLIP-IQA $\uparrow$}
&
{\scriptsize TOPIQ $\uparrow$}
&
{\scriptsize LIQE $\uparrow$}
\\

\midrule
Baseline
& 71.68 & 31.11 & 58.12 & 66.01 & 73.63 & 62.49 & 4.05 \\

IA$^3$~\cite{liu2022few}
& 73.95 & 30.58 & 59.47 & 67.18 & 74.44 & 64.55 & 4.31 \\

Houlsby~\cite{houlsby2019parameter}
& 74.47 & 30.85 & 59.65 & 67.77 & 75.08 & 66.34 & 4.41 \\

LoReFT~\cite{wu2024reft}
& 74.73 & 30.78 & \textbf{60.60} & 68.07 & 75.73 & 66.88 & 4.45 \\

Ch.-local LoRA~\cite{hu2022lora}
& 74.91
& 30.70
& 60.26
& 67.93
& 75.01
& 66.54
& 4.43
\\

\midrule
Param.-match LoRA~\cite{hu2022lora}
& \textbf{76.42}
& \textbf{26.87}
& 60.41
& 66.63
& 72.39
& 63.29
& \textbf{4.86}
\\
\rowcolor{OurColor}
\textbf{\ours (Ours)}
& 73.94 & 31.07 & 60.13 & \textbf{68.16} & \textbf{76.32} & \textbf{67.39} & 4.47 \\

\bottomrule
\end{tabular}
}

\vspace{-0.2cm}

\caption{
\textbf{Comparison with alternative adaptation mechanisms on DRealSR.} IA$^3$, Houlsby, LoReFT, channel-localized LoRA operate on the same top-$8$ channels as \ours, while parameter-matched LoRA uses a comparable trainable parameter budget.}

\label{tab:intervention_ablation}
\vspace{-0.55cm}

\end{table}

\begin{figure}[t]
    \centering
    \includegraphics[width=.98\linewidth]{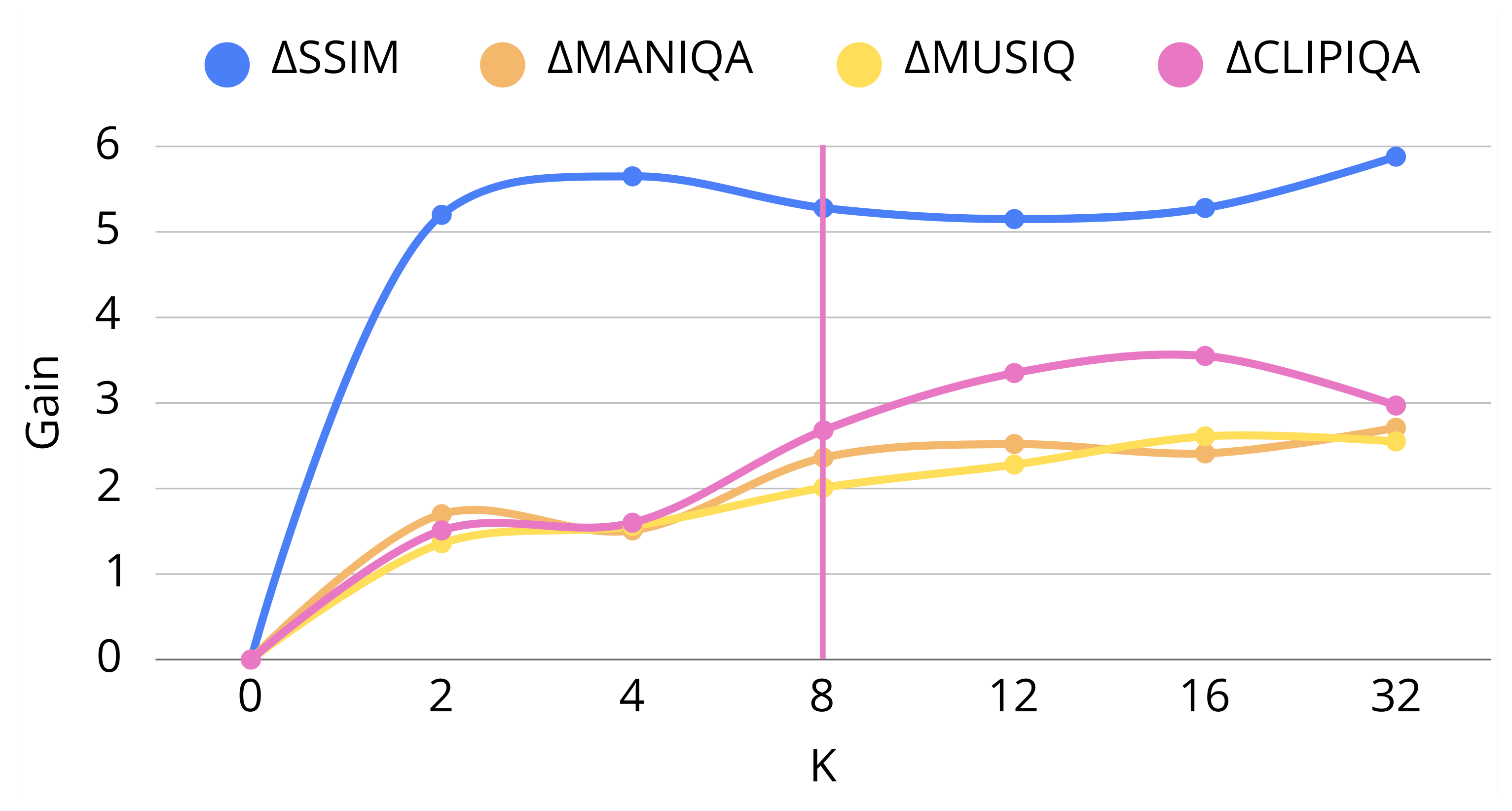}
    \vspace{-0.5cm}
    \caption{\textbf{Effect of the number of selected channels $K$.}}
    \label{fig:ablation_on_k}
    \vspace{-0.6cm}
\end{figure}

\tit{Effect of the number of selected channels}
Fig.~\ref{fig:ablation_on_k} analyzes the effect of varying the number of selected channels $K$. As shown, performance improves rapidly for small values of $K$, with most of the gain already recovered by $K=8$, making it a compact and effective operating point. Increasing $K$ further yields smaller and metric-dependent improvements, indicating diminishing returns as additional channels are included. In particular, MANIQA and MUSIQ improve gradually, while CLIP-IQA benefits up to intermediate values of $K$ and SSIM remains consistently high. These results are consistent with the analysis in Sec.~\ref{sec:analysis}, showing that a small subset of dominant channels is sufficient to exert substantial control over reconstruction quality.

\tit{Robustness to the perceptual objective}
Since LIQE is used both for training and evaluation, we test whether the observed improvements depend on optimizing this specific IQA metric. We retrain the predictor by replacing LIQE with MANIQA, MUSIQ, or TOPIQ, while keeping the remaining setup unchanged and adjusting only the corresponding loss weight to account for differences in numerical scale.
As shown in Table~\ref{tab:iqa_objective}, all objectives retain strong cross-metric performance, with gains consistently transferring to perceptual metrics not directly optimized during training. These results indicate that the perceptual gains of \ours are not specific to LIQE or explained by metric overfitting.

\tit{Efficiency and adaptation cost}
\ours leaves the pretrained backbone unchanged: both the DiT and the VAE remain frozen, and adaptation is carried entirely by a lightweight predictor with $274$K trainable parameters, fewer than the $361$K used by the parameter-matched LoRA baseline (cf. Table~\ref{tab:intervention_ablation}) and orders of magnitude below the backbones themselves. 
Phase~1 is also inexpensive, requiring only forward passes without gradient computation and storing per-channel statistics. On TSD-SR, the stability criterion typically terminates after roughly $400$ images ($\sim$25 mini-batches), taking about $8$ minutes on a single GPU. Phase~2 then trains for approximately $2.5$ hours on a single NVIDIA L40S GPU. 
At inference, the predictor runs once per image on the VAE latent already computed by the SR pipeline, while modulation consists only of affine transformations on the $K$ selected channels per stream and block, leaving the computational graph of the backbone unchanged.

\begin{table}[t]
\centering
\footnotesize
\setlength{\tabcolsep}{1.2pt}
\resizebox{\linewidth}{!}{
\begin{tabular}{
@{}
l
cc
ccccc
>{\columncolor{lightgray!50}}c
@{}
}

\toprule

&
\multicolumn{2}{c}{\textbf{Fidelity}}
&
\multicolumn{5}{c}{\textbf{Perceptual Quality}}
\\

\cmidrule(lr){2-3}
\cmidrule(lr){4-8}

&
{\scriptsize SSIM $\uparrow$}
&
{\scriptsize LPIPS $\downarrow$}
&
{\scriptsize MANIQA $\uparrow$}
&
{\scriptsize MUSIQ $\uparrow$}
&
{\scriptsize CLIP-IQA $\uparrow$}
&
{\scriptsize TOPIQ $\uparrow$}
&
{\scriptsize LIQE $\uparrow$}
\\

\midrule

Baseline
& 71.68 & 31.11 & 58.12 & 66.01 & 73.63 & 62.49 & 4.05 \\

Bottom
& 73.25
& 30.59
& 58.16
& 66.47
& 73.26
& 64.40
& 4.18
\\

Random
& 73.37
& \textbf{30.35}
& 58.84
& 66.77
& 73.78
& 63.34
& 4.21
\\

\rowcolor{OurColor}
\textbf{Top (Ours)}
& \textbf{73.94}
& 31.07
& \textbf{60.13}
& \textbf{68.16}
& \textbf{76.32}
& \textbf{67.39}
& \textbf{4.47}
\\

\bottomrule
\end{tabular}
}

\vspace{-0.2cm}

\caption{\textbf{Effect of channel selection strategy.} Top, random, and bottom selections are evaluated using $K=8$.}
\label{tab:selection_ablation}
\vspace{-0.25cm}
\end{table}

\begin{table}[t]
\centering
\footnotesize
\setlength{\tabcolsep}{1.5pt}

\resizebox{\linewidth}{!}{
\begin{tabular}{
@{}
l
cc
ccccc
@{}
}

\toprule

&
\multicolumn{2}{c}{\textbf{Fidelity}}
&
\multicolumn{5}{c}{\textbf{Perceptual Quality}}
\\

\cmidrule(lr){2-3}
\cmidrule(lr){4-8}

&
{\scriptsize SSIM $\uparrow$}
&
{\scriptsize LPIPS $\downarrow$}
&
{\scriptsize MANIQA $\uparrow$}
&
{\scriptsize MUSIQ $\uparrow$}
&
{\scriptsize CLIP-IQA $\uparrow$}
&
{\scriptsize TOPIQ $\uparrow$}
&
{\scriptsize LIQE $\uparrow$}
\\

\midrule
\rowcolor{OurColor}
\textbf{LIQE (Ours)} & 73.94 & 31.07 & \textbf{60.13} & \textbf{68.16} & 76.32 & 67.39 & 4.47 \\
MANIQA & 71.86 & 32.73 & 59.91 & 68.12 & \textbf{79.06} & 69.40 & \textbf{4.50} \\
MUSIQ & \textbf{74.75} & \textbf{29.28} & 59.60 & 67.32 & 72.61 & 64.28 & 4.22 \\
TOPIQ & 73.43 & 31.02 & 59.42 & 67.97 & 75.34 & \textbf{69.64} & 4.37 \\

\bottomrule
\end{tabular}
}

\vspace{-0.2cm}

\caption{\textbf{Robustness to the perceptual training objective.} LIQE, MANIQA, MUSIQ, and TOPIQ are used with loss weight $\alpha=0.1$, $0.5$, $0.005$, and $0.5$, respectively.}

\label{tab:iqa_objective}
\vspace{-0.3cm}

\end{table}

\section{Conclusion}
\label{sec:conclusion}

We investigated magnitude-dominant activation channels as a compact interface for frozen DiT SR models. We show that they form a sparse, intervention-sensitive subspace. Based on this insight, we introduced \ours, an input-conditioned controller that modulates selected channels while keeping the backbone frozen. Across three SR backbones and multiple benchmarks, \ours improves fidelity and perceptual quality, highlighting sparse representation-space intervention as an effective alternative to weight-space adaptation.

{
    \small
    \bibliographystyle{ieeenat_fullname}
    \bibliography{bibliography}

@string{cvpr     = {CVPR}}

@string{cvprw    = {CVPR Workshops}}

@string{nips     = {NeurIPS}}

@string{iccv     = {ICCV}}

@string{eccv     = {ECCV}}

@string{eccvw    = {ECCV Workshops}}

@string{iclr     = {ICLR}}

@string{icml     = {ICML}}

@string{icpr     = {ICPR}}

@string{aaai     = {AAAI}}

@string{acl      = {ACL}}

@string{colm     = {COLM}}

@string{ieeetpami  = {IEEE TPAMI}}

@string{ieeetip  = {IEEE TIP}}

@string{ijcv     = {IJCV}}

@string{ieeetpami  = {IEEE Trans. PAMI}}

@string{ieeetip    = {IEEE Trans. Image Processing}}

@article{turri2026few,
  title={{Few Channels Draw The Whole Picture: Revealing Massive Activations in Diffusion Transformers}},
  author={Turri, Evelyn and Bucciarelli, Davide and Sarto, Sara and Baraldi, Lorenzo and Cornia, Marcella},
  journal={arXiv preprint arXiv:2605.13974},
  year={2026}
}

@article{topiq,
  title={{TOPIQ: A Top-Down Approach from Semantics to Distortions for Image Quality Assessment}},
  author={Chen, Chaofeng and Mo, Jiadi and Hou, Jingwen and Wu, Haoning and Liao, Liang and Sun, Wenxiu and Yan, Qiong and Lin, Weisi},
  journal=ieeetip,
  volume={33},
  pages={2404--2418},
  year={2024}
}

@inproceedings{sun2025pixel,
  title={{Pixel-level and Semantic-level Adjustable Super-resolution: A Dual-LoRA Approach}},
  author={Sun, Lingchen and Wu, Rongyuan and Ma, Zhiyuan and Liu, Shuaizheng and Yi, Qiaosi and Zhang, Lei},
  booktitle=cvpr,
  year={2025}
}

@inproceedings{wang2021real,
  title={{Real-ESRGAN: Training Real-World Blind Super-Resolution With Pure Synthetic Data}},
  author={Wang, Xintao and Xie, Liangbin and Dong, Chao and Shan, Ying},
  booktitle=iccv,
  year={2021}
}

@inproceedings{agustsson2017ntire,
  title={{NTIRE 2017 Challenge on Single Image Super-Resolution: Dataset and Study}},
  author={Agustsson, Eirikur and Timofte, Radu},
  booktitle=cvprw,
  year={2017}
}

@inproceedings{cai2019toward,
  title={{Toward Real-World Single Image Super-Resolution: A New Benchmark and a New Model}},
  author={Cai, Jianrui and Zeng, Hui and Yong, Hongwei and Cao, Zisheng and Zhang, Lei},
  booktitle=iccv,
  year={2019}
}

@inproceedings{wei2020component,
  title={{Component Divide-and-Conquer for Real-World Image Super-Resolution}},
  author={Wei, Pengxu and Xie, Ziwei and Lu, Hannan and Zhan, Zongyuan and Ye, Qixiang and Zuo, Wangmeng and Lin, Liang},
  booktitle=eccv,
  year={2020},
}

@inproceedings{zhang2018unreasonable,
  title={{The Unreasonable Effectiveness of Deep Features as a Perceptual Metric}},
  author={Zhang, Richard and Isola, Phillip and Efros, Alexei A and Shechtman, Eli and Wang, Oliver},
  booktitle=cvpr,
  year={2018}
}

@inproceedings{dong2014learning,
  title={{Learning a Deep Convolutional Network for Image Super-Resolution}},
  author={Dong, Chao and Loy, Chen Change and He, Kaiming and Tang, Xiaoou},
  booktitle=eccv,
  year={2014},
}

@inproceedings{lim2017enhanced,
  title={{Enhanced Deep Residual Networks for Single Image Super-Resolution}},
  author={Lim, Bee and Son, Sanghyun and Kim, Heewon and Nah, Seungjun and Mu Lee, Kyoung},
  booktitle=cvprw,
  year={2017}
}

@inproceedings{zhang2021designing,
  title={{Designing a Practical Degradation Model for Deep Blind Image Super-Resolution}},
  author={Zhang, Kai and Liang, Jingyun and Van Gool, Luc and Timofte, Radu},
  booktitle=iccv,
  year={2021}
}

@article{wang2024exploiting,
  title={{Exploiting Diffusion Prior for Real-World Image Super-Resolution}},
  author={Wang, Jianyi and Yue, Zongsheng and Zhou, Shangchen and Chan, Kelvin CK and Loy, Chen Change},
  journal=ijcv,
  volume={132},
  number={12},
  pages={5929--5949},
  year={2024},
}

@inproceedings{yang2024pixel,
  title={{Pixel-Aware Stable Diffusion for Realistic Image Super-Resolution and Personalized Stylization}},
  author={Yang, Tao and Wu, Rongyuan and Ren, Peiran and Xie, Xuansong and Zhang, Lei},
  booktitle=eccv,
  year={2024},
}

@inproceedings{wang2024sinsr,
  title={{SinSR: Diffusion-Based Image Super-Resolution in a Single Step}},
  author={Wang, Yufei and Yang, Wenhan and Chen, Xinyuan and Wang, Yaohui and Guo, Lanqing and Chau, Lap-Pui and Liu, Ziwei and Qiao, Yu and Kot, Alex C and Wen, Bihan},
  booktitle=cvpr,
  year={2024}
}

@inproceedings{wu2024seesr,
  title={{SeeSR: Towards Semantics-Aware Real-World Image Super-Resolution}},
  author={Wu, Rongyuan and Yang, Tao and Sun, Lingchen and Zhang, Zhengqiang and Li, Shuai and Zhang, Lei},
  booktitle=cvpr,
  year={2024}
}

@inproceedings{yu2024scaling,
  title={{Scaling Up to Excellence: Practicing Model Scaling for Photo-Realistic Image Restoration In the Wild}},
  author={Yu, Fanghua and Gu, Jinjin and Li, Zheyuan and Hu, Jinfan and Kong, Xiangtao and Wang, Xintao and He, Jingwen and Qiao, Yu and Dong, Chao},
  booktitle=cvpr,
  year={2024}
}

@article{zhang2024degradation,
  title={{Degradation-Guided One-Step Image Super-Resolution with Diffusion Priors}},
  author={Zhang, Aiping and Yue, Zongsheng and Pei, Renjing and Ren, Wenqi and Cao, Xiaochun},
  journal={arXiv preprint},
  year={2024}
}

@inproceedings{hu2022lora,
  title={{LoRA: Low-Rank Adaptation of Large Language Models}},
  author={Hu, Edward J and Shen, Yelong and Wallis, Phillip and Allen-Zhu, Zeyuan and Li, Yuanzhi and Wang, Shean and Wang, Lu and Chen, Weizhu and others},
  booktitle=iclr,
  year={2022}
}

@article{gao2026teasr,
  title={{TEASR: Training-Efficient Any-Step Diffusion Transformer for Real-World Image Super-Resolution}},
  author={Gao, Xiang and Zhu, Chenxin and Fang, Yushun and Hu, Qiang and Zhang, Xiaoyun},
  journal={arXiv preprint arXiv:2606.16188},
  year={2026}
}

@inproceedings{houlsby2019parameter,
  title={{Parameter-efficient transfer learning for NLP}},
  author={Houlsby, Neil and Giurgiu, Andrei and Jastrzebski, Stanislaw and Morrone, Bruna and De Laroussilhe, Quentin and Gesmundo, Andrea and Attariyan, Mona and Gelly, Sylvain},
  booktitle=icml,
  year={2019},
}

@inproceedings{wu2024reft,
  title={{ReFT: Representation Finetuning for Language Models}},
  author={Wu, Zhengxuan and Arora, Aryaman and Wang, Zheng and Geiger, Atticus and Jurafsky, Dan and Manning, Christopher D and Potts, Christopher},
  booktitle=nips,
  year={2024}
}

@inproceedings{liu2022few,
  title={{Few-shot parameter-efficient fine-tuning is better and cheaper than in-context learning}},
  author={Liu, Haokun and Tam, Derek and Muqeeth, Mohammed and Mohta, Jay and Huang, Tenghao and Bansal, Mohit and Raffel, Colin A},
  booktitle=nips,
  year={2022}
}

@inproceedings{gan2025unleashing,
  title={{Unleashing Diffusion Transformers for Visual Correspondence by Modulating Massive Activations}},
  author={Gan, Chaofan and Tu, Yuanpeng and Chen, Xi and Chen, Tieyuan and Li, Yuxi and Harandi, Mehrtash and Lin, Weiyao},
  booktitle=nips,
  year={2025}
}

@inproceedings{wu2024one,
  title={{One-Step Effective Diffusion Network for Real-World Image Super-Resolution}},
  author={Wu, Rongyuan and Sun, Lingchen and Ma, Zhiyuan and Zhang, Lei},
  booktitle=nips,
  year={2024}
}

@article{he2026one,
  title={{One step diffusion-based super-resolution with time-aware distillation}},
  author={He, Xiao and Tang, Haoao and Tu, Zhijun and Zhang, Junchao and Cheng, Kun and Chen, Hanting and Guo, Yong and Zhu, Mingrui and Hu, Jie and Wang, Nannan and others},
  journal=ieeetip,
  year={2026},
}

@article{tai2026addsr,
  title={{AddSR: Accelerating diffusion-based blind super-resolution with adversarial diffusion distillation}},
  author={Tai, Ying and Xie, Rui and Zhao, Chen and Zhang, Kai and Zhang, Zhenyu and Zhou, Jun and Yang, Jian},
  journal={Pattern Recognition},
  volume = {175},
  pages = {113012},
  year={2026}
}

@inproceedings{cheng2025effective,
  title={{Effective diffusion transformer architecture for image super-resolution}},
  author={Cheng, Kun and Yu, Lei and Tu, Zhijun and He, Xiao and Chen, Liyu and Guo, Yong and Zhu, Mingrui and Wang, Nannan and Gao, Xinbo and Hu, Jie},
  booktitle=aaai,
  year={2025}
}

@inproceedings{duan2025dit4sr,
  title={{DiT4SR: Taming Diffusion Transformer for Real-World Image Super-Resolution}},
  author={Duan, Zheng-Peng and Zhang, Jiawei and Jin, Xin and Zhang, Ziheng and Xiong, Zheng and Zou, Dongqing and Ren, Jimmy S and Guo, Chunle and Li, Chongyi},
  booktitle=iccv,
  year={2025}
}

@inproceedings{dong2025tsd,
  title={{TSD-SR: One-Step Diffusion with Target Score Distillation for Real-World Image Super-Resolution}},
  author={Dong, Linwei and Fan, Qingnan and Guo, Yihong and Wang, Zhonghao and Zhang, Qi and Chen, Jinwei and Luo, Yawei and Zou, Changqing},
  booktitle=cvpr,
  year={2025}
}

@inproceedings{wang2018esrgan,
  title={{Esrgan: Enhanced super-resolution generative adversarial networks}},
  author={Wang, Xintao and Yu, Ke and Wu, Shixiang and Gu, Jinjin and Liu, Yihao and Dong, Chao and Qiao, Yu and Change Loy, Chen},
  booktitle=eccvw,
  year={2018}
}

@inproceedings{li2025one,
  title={{One Diffusion Step to Real-World Super-Resolution via Flow Trajectory Distillation}},
  author={Li, Jianze and Cao, Jiezhang and Guo, Yong and Li, Wenbo and Zhang, Yulun},
  booktitle=icml,
  year={2025}
}

@inproceedings{lin2024diffbir,
  title={{DiffBIR: Toward Blind Image Restoration with Generative Diffusion Prior}},
  author={Lin, Xinqi and He, Jingwen and Chen, Ziyan and Lyu, Zhaoyang and Dai, Bo and Yu, Fanghua and Qiao, Yu and Ouyang, Wanli and Dong, Chao},
  booktitle=eccv,
  year={2024},
}

@inproceedings{fdoronzio2026gramsr,
  title={{GramSR: Visual Feature Conditioning for Diffusion-Based Super-Resolution}},
  author={D'Oronzio, Fabio and Putamorsi, Federico and Zini, Leonardo and Cornia, Marcella and Baraldi, Lorenzo},
  booktitle=icpr,
  year={2026}
}

@inproceedings{yang2022maniqa,
  title={{Maniqa: Multi-dimension attention network for no-reference image quality assessment}},
  author={Yang, Sidi and Wu, Tianhe and Shi, Shuwei and Lao, Shanshan and Gong, Yuan and Cao, Mingdeng and Wang, Jiahao and Yang, Yujiu},
  booktitle=cvpr,
  year={2022}
}

@inproceedings{ke2021musiq,
  title={{MUSIQ: Multi-scale Image Quality Transformer}},
  author={Ke, Junjie and Wang, Qifei and Wang, Yilin and Milanfar, Peyman and Yang, Feng},
  booktitle=iccv,
  year={2021}
}

@inproceedings{wang2023exploring,
  title={{Exploring clip for assessing the look and feel of images}},
  author={Wang, Jianyi and Chan, Kelvin CK and Loy, Chen Change},
  booktitle=aaai,
  year={2023}
}

@inproceedings{zhang2023blind,
  title={{Blind image quality assessment via vision-language correspondence: A multitask learning perspective}},
  author={Zhang, Weixia and Zhai, Guangtao and Wei, Ying and Yang, Xiaokang and Ma, Kede},
  booktitle=cvpr,
  year={2023}
}

@article{labs2025flux,
  title={{FLUX.1 Kontext: Flow Matching for In-Context Image Generation and Editing in Latent Space}},
  author={Labs, Black Forest and Batifol, Stephen and Blattmann, Andreas and Boesel, Frederic and Consul, Saksham and Diagne, Cyril and Dockhorn, Tim and English, Jack and English, Zion and Esser, Patrick and others},
  journal={arXiv preprint},
  year={2025}
}

@inproceedings{zhao2024vidit,
  title={{ViDiT-Q: Efficient and Accurate Quantization of Diffusion Transformers for Image and Video Generation}},
  author={Zhao, Tianchen and Fang, Tongcheng and Huang, Haofeng and Liu, Enshu and Wan, Rui and Soedarmadji, Widyadewi and Li, Shiyao and Lin, Zinan and Dai, Guohao and Yan, Shengen and others},
  booktitle=iclr,
  year={2025}
}

@inproceedings{peebles2023scalable,
  title={Scalable diffusion models with transformers},
  author={Peebles, William and Xie, Saining},
  booktitle=iccv,
  year={2023}
}

@inproceedings{esser2024scaling,
  title={{Scaling Rectified Flow Transformers for High-Resolution Image Synthesis}},
  author={Esser, Patrick and Kulal, Sumith and Blattmann, Andreas and Entezari, Rahim and M{\"u}ller, Jonas and Saini, Harry and Levi, Yam and Lorenz, Dominik and Sauer, Axel and Boesel, Frederic and others},
  booktitle=icml,
  year={2024}
}

@inproceedings{sun2024massive,
  title={{Massive Activations in Large Language Models}},
  author={Sun, Mingjie and Chen, Xinlei and Kolter, J Zico and Liu, Zhuang},
  booktitle=colm,
  year={2024}
}

@inproceedings{yang2024denoising,
  title={{Denoising Vision Transformers}},
  author={Yang, Jiawei and Luo, Katie Z and Li, Jiefeng and Deng, Congyue and Guibas, Leonidas and Krishnan, Dilip and Weinberger, Kilian Q and Tian, Yonglong and Wang, Yue},
  booktitle=eccv,
  year={2024},
}

@inproceedings{zhao2023unveiling,
  title={{Unveiling Linguistic Regions in Large Language Models}},
  author={Zhao, Jun and Zhang, Zhihao and Ma, Yide and Zhang, Qi and Gui, Tao and Gao, Luhui and Huang, Xuanjing},
  booktitle=acl,
  year={2024}
}

@inproceedings{fang2025tinyfusion,
  title={{TinyFusion: Diffusion Transformers Learned Shallow}},
  author={Fang, Gongfan and Li, Kunjun and Ma, Xinyin and Wang, Xinchao},
  booktitle=cvpr,
  year={2025}
}

@inproceedings{Blau_2018_CVPR,
  title={{The Perception-Distortion Tradeoff}},
  author={Blau, Yochai and Michaeli, Tomer},
  booktitle=cvpr,
  year={2018}
}

@article{dong2015image,
  title={{Image super-resolution using deep convolutional networks}},
  author={Dong, Chao and Loy, Chen Change and He, Kaiming and Tang, Xiaoou},
  journal=ieeetpami,
  volume={38},
  number={2},
  pages={295--307},
  year={2015},
}

@inproceedings{ledig2017photo,
  title={{Photo-realistic single image super-resolution using a generative adversarial network}},
  author={Ledig, Christian and Theis, Lucas and Husz{\'a}r, Ferenc and Caballero, Jose and Cunningham, Andrew and Acosta, Alejandro and Aitken, Andrew and Tejani, Alykhan and Totz, Johannes and Wang, Zehan and others},
  booktitle=cvpr,
  year={2017}
}

@inproceedings{Rombach_2022_CVPR,
    author    = {Rombach, Robin and Blattmann, Andreas and Lorenz, Dominik and Esser, Patrick and Ommer, Bj\"orn},
    title     = {{High-Resolution Image Synthesis With Latent Diffusion Models}},
    booktitle = cvpr,
    year      = {2022}
}

@article{lipman2024flowmatchingguidecode,
      title={{Flow Matching Guide and Code}}, 
      author={Yaron Lipman and Marton Havasi and Peter Holderrieth and Neta Shaul and Matt Le and Brian Karrer and Ricky T. Q. Chen and David Lopez-Paz and Heli Ben-Hamu and Itai Gat},
      year={2024},
      journal={arXiv preprint}
}

@inproceedings{ho2020denoising,
  title={Denoising diffusion probabilistic models},
  author={Ho, Jonathan and Jain, Ajay and Abbeel, Pieter},
  booktitle=nips,
  year={2020}
}

@inproceedings{liu2022flow,
  title={{Flow Straight and Fast: Learning to Generate and Transfer Data with Rectified Flow}},
  author={Liu, Xingchao and Gong, Chengyue and Liu, Qiang},
  booktitle=iclr,
  year={2023}
}

@article{wu2025dp2osr,
  title     = {{DP$^2$O-SR: Direct Perceptual Preference Optimization for Real-World Image Super-Resolution}},
  author    = {Wu, Rongyuan and Sun, Lingchen and Zhang, Zhengqiang and Wang, Shihao and Wu, Tianhe and Yi, Qiaosi and Li, Shuai and Zhang, Lei},
  journal={arXiv preprint},
  year      = {2025}
}

@inproceedings{yi2025fine,
  title={{Fine-Structure Preserved Real-World Image Super-Resolution Via Transfer VAE Training}},
  author={Yi, Qiaosi and Li, Shuai and Wu, Rongyuan and Sun, Lingchen and Wu, Yuhui and Zhang, Lei},
  booktitle=iccv,
  year={2025}
}

@article{sun2025pocketsr,
  title={{PocketSR: The Super-Resolution Expert in Your Pocket Mobiles}},
  author={Sun, Haoze and Jiang, Linfeng and Li, Fan and Pei, Renjing and Wang, Zhixin and Guo, Yong and Xu, Jiaqi and Chen, Haoyu and Han, Jin and Song, Fenglong and others},
  journal={arXiv preprint},
  year={2025}
}

@inproceedings{
darcet2024vision,
title={{Vision Transformers Need Registers}},
author={Timoth{\'e}e Darcet and Maxime Oquab and Julien Mairal and Piotr Bojanowski},
booktitle=iclr,
year={2024}
}

@inproceedings{gan2025massive,
  title={{Massive Activations are the Key to Local Detail Synthesis in Diffusion Transformers}},
  author={Gan, Chaofan and Zhao, Zicheng and Tu, Yuanpeng and Chen, Xi and Qin, Ziran and Chen, Tieyuan and Harandi, Mehrtash and Lin, Weiyao},
  booktitle=iclr,
  year={2026}
}

@article{10.1109/TIT.2009.2027527, 
author = {Hurley, Niall and Rickard, Scott}, 
title = {Comparing measures of sparsity}, 
  volume={55},
  number={10},
  pages={4723--4741},
  year={2009},
journal = {IEEE Trans. Inf. Theor.}
}

@ARTICLE{1284395,
  author={Zhou Wang and Bovik, A.C. and Sheikh, H.R. and Simoncelli, E.P.},
  journal=ieeetip, 
  title={Image quality assessment: from error visibility to structural similarity}, 
  year={2004},
}

@inproceedings{kingma2014adam,
  title={{Adam: A Method for Stochastic Optimization}},
  author={Kingma, Diederik P and Ba, Jimmy},
  booktitle=iclr,
  year={2015}
}

@inproceedings{austin2021structured,
title={Structured Denoising Diffusion Models in Discrete State-Spaces},
author={Jacob Austin and Daniel D. Johnson and Jonathan Ho and Daniel Tarlow and Rianne van den Berg},
booktitle=nips,
year={2021}
}

@article{bertolani2026diffusion,
  title={Diffusion Language Models: An Experimental Analysis},
  author={Bertolani, Thomas and Bucciarelli, Davide and Zini, Leonardo and Cornia, Marcella and Baraldi, Lorenzo},
  journal={arXiv preprint},
  year={2026}
}
}

\newpage
\appendix

\maketitlesupplementary

\section{Additional Implementation Details}

\subsection{Computational Resources}

Phase~1 requires a single partial forward pass over the optimization dataset with the SR backbone frozen and no gradient computation. For TSD-SR~\cite{dong2025tsd}, early stopping typically occurs after processing $\sim$400 images ($\sim$25 mini-batches of size 16), taking approximately 8 minutes on a single GPU; for TEASR~\cite{gao2026teasr}, 300 images suffice and are processed in about 5 minutes. Memory overhead during Phase~1 is negligible, as only per-channel mean activations are stored.

Phase~2 is run on a single NVIDIA L40S GPU with 48 GB of memory. We use an effective batch size of 16 across all models, achieved via gradient accumulation. The same setup is used for all baselines, including LoRA~\cite{hu2022lora} and the other parameter-efficient variants. For TEASR, whose backbone is a 20B-parameter DiT, the frozen weights are stored in FP8 ($\sim$20.5 GB) with bf16 compute, keeping the peak footprint at $\sim$26 GB and allowing the whole pipeline to run on the same single 48 GB GPU.

Training times vary across backbones: TSD-SR requires approximately 2.5 hours, TEASR around 15 hours, and DiT4SR~\cite{duan2025dit4sr} up to 36 hours due to its multi-step formulation. Inference on DIV2K~\cite{agustsson2017ntire} takes approximately 30 minutes for TSD-SR, $\sim$1 hour for TEASR, and $\sim$7 hours for DiT4SR. Evaluation on RealSR~\cite{cai2019toward} and DRealSR~\cite{wei2020component} is substantially cheaper, taking about 5 minutes for TSD-SR, $\sim$10 minutes for TEASR, and scaling proportionally for the other backbones. Adaptation therefore introduces only a marginal overhead over the cost of running the frozen backbone itself.

\subsection{Multi-Seed Stability Analysis}

To assess the stability of the proposed approach, all experiments on TSD-SR are repeated with three different random seeds on DRealSR, and the results reported in the main paper are averages across runs.

The overall mean standard deviation across metrics and configurations is $0.0021$ after normalizing the metric scales, indicating low cross-seed variability. At the metric level, MUSIQ~\cite{ke2021musiq} exhibits the lowest cross-seed variability ($\sigma = 0.0013$), followed closely by MANIQA~\cite{yang2022maniqa} and LPIPS~\cite{zhang2018unreasonable} (both $\sigma = 0.0016$), while CLIP-IQA~\cite{wang2023exploring} ($\sigma = 0.0028$) and LIQE~\cite{zhang2023blind} ($\sigma = 0.0034$) are the most variable. Nevertheless, the observed variations remain small relative to the performance differences between configurations and do not alter their overall ranking.
At the level of selection strategies, bottom-$K$ selection exhibits the lowest variability ($\sigma = 0.0014$), followed by top-$K$ ($\sigma = 0.0022$) and random selection ($\sigma = 0.0027$). The slightly higher variability of random selection can be attributed to the additional stochasticity introduced by drawing the modulated channels at random, whereas top-$K$ and bottom-$K$ rely on deterministic activation-based criteria. Despite this difference, variability remains small across all selection strategies, and the relative performance trends are preserved across seeds.

\subsection{Pseudocode}

Algorithm~\ref{alg:phase1} and Algorithm~\ref{alg:phase2} provide detailed pseudocode for Phase~1 (online channel selection) and Phase~2 (image-conditioned predictor training), respectively. Together, they summarize the full pipeline, including the estimation of channel importance, the construction of the modulation subspace, and the training procedure used to learn input-conditioned affine parameters. Equation numbers refer to the main paper.

\begin{algorithm}[!t]
\caption{Phase~1 -- Online Channel Selection via EMA}
\label{alg:phase1}
\begin{algorithmic}[1]
\Require Pretrained SR backbone $\mathcal{F}$, optimization set $\mathcal{D}$,
         EMA decay $\lambda$, window length $W$, number of window
         transitions $P$, stability threshold $\tau$, budget $K$
\Ensure  Selected channel sets $\mathcal{S} = \{(\ell, s, \mathcal{I}^{(\ell,s)})\}$
\State Initialize $m_{0}^{(\ell,s)} \leftarrow \mathbf{0}$ for all blocks $\ell$ and streams $s$
\State $t \leftarrow 0$,\quad $w \leftarrow 0$
\For{each mini-batch $\mathcal{B} \subset \mathcal{D}$}
    \State $t \leftarrow t + 1$
    \State Run a forward pass of $\mathcal{F}$ on $\mathcal{B}$ without gradients; collect $a_{t}^{(\ell,s)}$ (Eq.~1) for all $(\ell, s)$
    \State $m_{t}^{(\ell,s)} \leftarrow \lambda\, m_{t-1}^{(\ell,s)} + (1-\lambda)\,a_{t}^{(\ell,s)}$ \quad (Eq.~3)
    \If{$t \bmod W = 0$} \Comment{snapshot the ranking every $W$ steps}
        \State $w \leftarrow w + 1$
        \State $\mathcal{I}_{w}^{(\ell,s)} \leftarrow \mathrm{top}$-$K$ indices of $m_{t}^{(\ell,s)}$ for all $(\ell, s)$
        \If{$w > P$}
            \State $J^{(\ell,s)} \leftarrow \dfrac{1}{P}\displaystyle\sum_{p=0}^{P-1}
                   \frac{|\mathcal{I}_{w-p}^{(\ell,s)} \cap \mathcal{I}_{w-p-1}^{(\ell,s)}|}
                        {|\mathcal{I}_{w-p}^{(\ell,s)} \cup \mathcal{I}_{w-p-1}^{(\ell,s)}|}$ \quad (Eq.~5)
            \If{$J^{(\ell,s)} \geq \tau$ for all $(\ell, s)$}
                \State \textbf{break} \Comment{ranking has converged}
            \EndIf
        \EndIf
    \EndIf
\EndFor
\State $\mathcal{S} \leftarrow \{(\ell, s, \mathcal{I}_{w}^{(\ell,s)})\}$ for all $(\ell, s)$ \quad (Eq.~6)
\State \Return $\mathcal{S}$
\end{algorithmic}
\end{algorithm}

\begin{algorithm}[!t]
\caption{Phase~2 -- Image-Conditioned Predictor Training}
\label{alg:phase2}
\begin{algorithmic}[1]
\Require optimization pairs $\mathcal{D}_{opt}$, validation pairs $\mathcal{D}_{val}$, selected channels $\mathcal{S}$ from Phase~1, frozen SR backbone, frozen VAE, modulation bounds $[\gamma_{\min},\gamma_{\max}]$ and $[\beta_{\min},\beta_{\max}]$, loss weights $\alpha_{\mathrm{LIQE}}, \alpha_{\mathrm{TV}}$
\Ensure trained predictor $f_\theta$

\State Prepare optimization and validation samples from $\mathcal{D}_{opt}$ and $\mathcal{D}_{val}$
\State Encode one LR sample with the frozen VAE and compute pooled latent feature dimension $d_z$
\State Build the transformation layout from $\mathcal{S}$ by assigning an affine pair $(\gamma,\beta)$ to each selected channel, block, and stream, giving $\mathrm{dim}_{\text{out}} = 2 \sum_{\ell,s} |\mathcal{I}^{(\ell,s)}|$ \quad (Eq.~9)
\State Initialize predictor $f_\theta: \mathbb{R}^{d_z} \rightarrow \mathbb{R}^{\mathrm{dim}_{\text{out}}}$ and optimizer
\State $best \leftarrow \infty$

\For{$t = 1$ to $T$}
    \State Sample a mini-batch $(x,y)$ from $\mathcal{D}_{opt}$
    \State $z \leftarrow \mathrm{VAE}(x)$
    \State $u \leftarrow \operatorname{AvgPool}(z)$
    \State $(\bar{\gamma},\bar{\beta}) \leftarrow \mathrm{Decode}(f_\theta(u), \mathcal{S})$ \quad (Eq.~8)
    \State $\gamma \leftarrow \gamma_{\min} + (\gamma_{\max}-\gamma_{\min})\,\sigma(\bar{\gamma})$;\quad
           $\beta \leftarrow \beta_{\min} + (\beta_{\max}-\beta_{\min})\,\sigma(\bar{\beta})$ \quad (Eq.~10)
    \State $\hat{y} \leftarrow \mathrm{SR\text{-}Model}(x;\gamma,\beta)$ \Comment{backbone frozen, gradients flow to $f_\theta$}
    \State Convert $\hat{y}$ to image space in $[0,1]$
    \State $\mathcal{L} \leftarrow
    \mathcal{L}_{\mathrm{LPIPS}}(\hat{y},y)
    - \alpha_{\mathrm{LIQE}}\mathrm{LIQE}(\hat{y})
    + \alpha_{\mathrm{TV}}\mathcal{L}_{\mathrm{TV}}(\hat{y})$ \quad (Eq.~11)
    \State Update $\theta$ by backpropagating $\mathcal{L}$

    \If{validation is performed}
        \State $\mathcal{L}_{val} \leftarrow$ validation loss on $\mathcal{D}_{val}$
        \If{$\mathcal{L}_{val} < best$}
            \State $best \leftarrow \mathcal{L}_{val}$; save checkpoint
        \EndIf
    \EndIf
\EndFor

\State \Return best predictor checkpoint
\end{algorithmic}
\end{algorithm}

\section{Extended Activation Analysis}

\begin{figure*}[!t]
    \centering
    \includegraphics[width=0.99\linewidth]{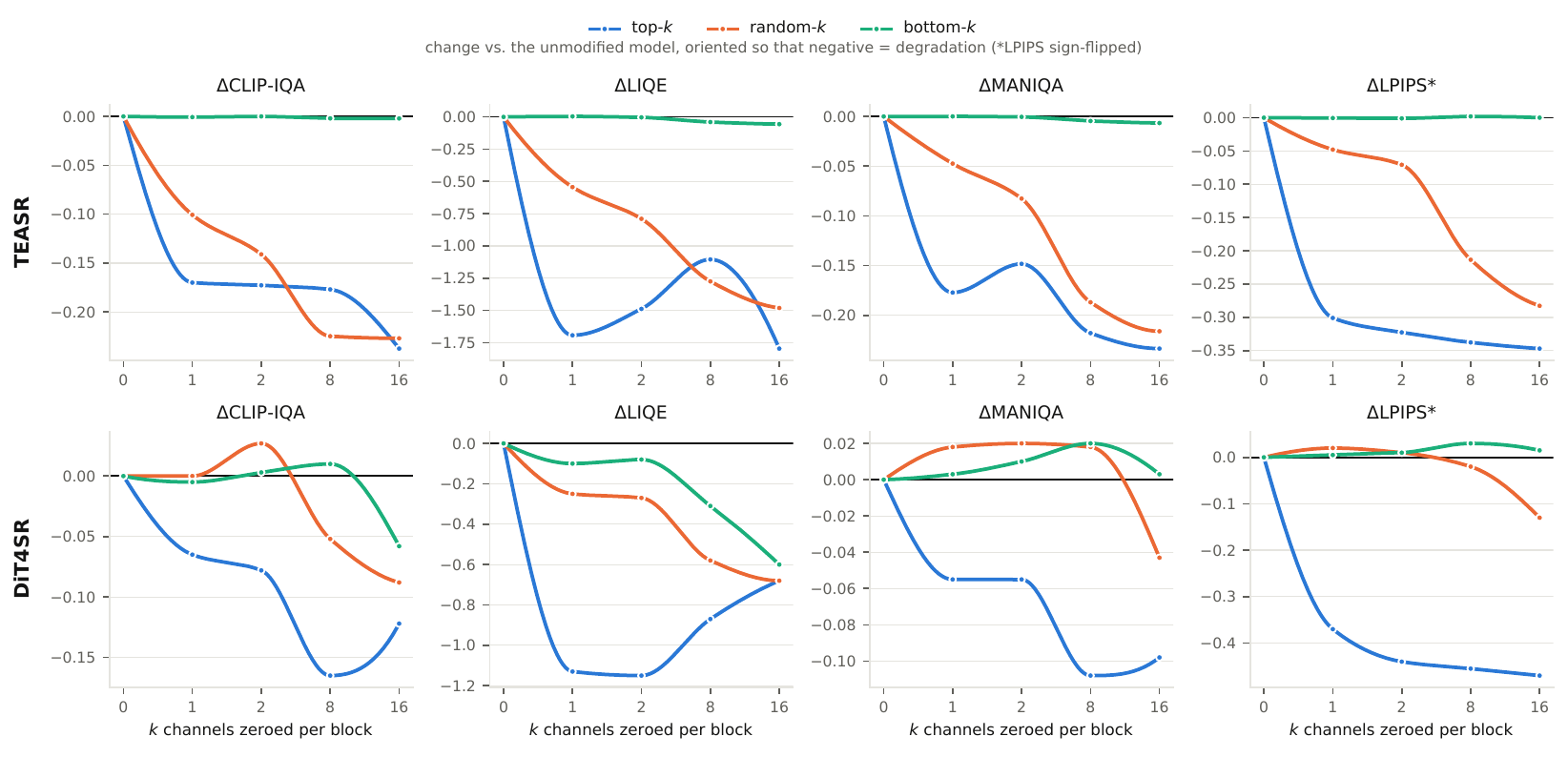}
    \vspace{-0.3cm}
    \caption{\textbf{Effect of zero ablation on TEASR (top) and DiT4SR (bottom).} Performance degradation is reported across all metrics when zeroing top-, random-, and bottom-ranked channels, highlighting the sensitivity of both models to their dominant activations. Curves report the change with respect to the unmodified model, oriented so that negative values denote degradation; LPIPS is therefore sign-flipped.
}
    \label{fig:zero_ablation_plot_overall}
    \vspace{-0.2cm}
\end{figure*}

\begin{figure*}[!t]
    \centering
    \includegraphics[width=0.99\linewidth]{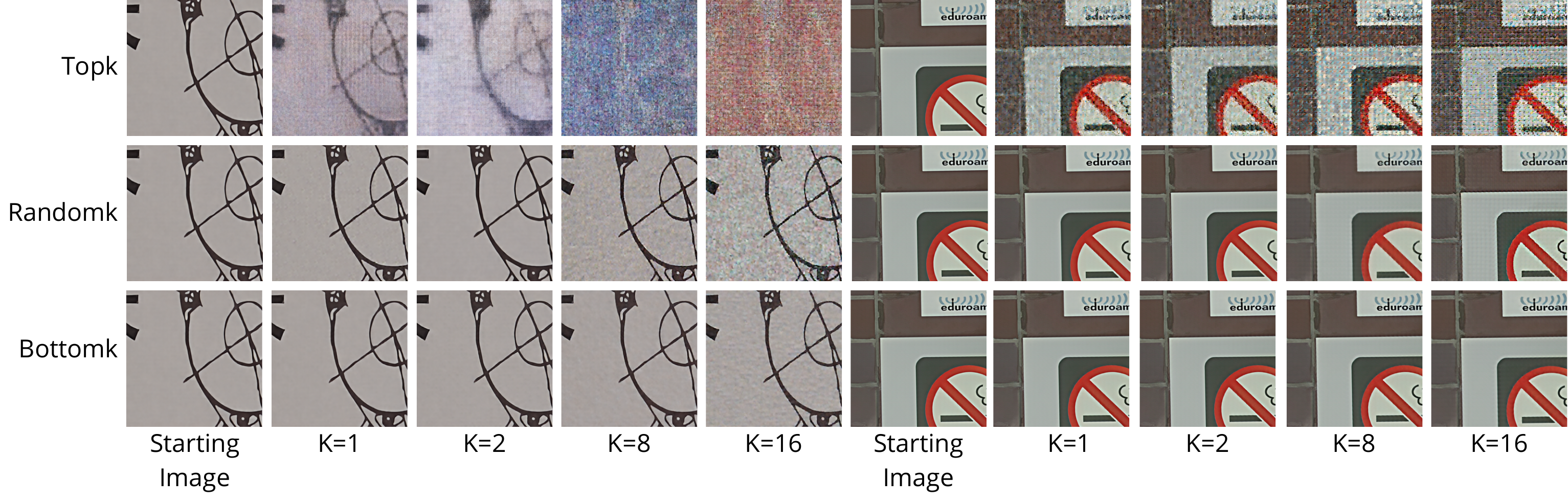}
    \vspace{-0.3cm}
    \caption{\textbf{Qualitative effect of zero ablation on DiT4SR (left) and TSD-SR (right).} For increasing values of $K$, zeroing top-ranked channels progressively destroys fine detail and structure, while random and bottom-ranked largely unaffect the reconstruction.}
    \label{fig:qualitative_disruption}
    \vspace{-0.55cm}
\end{figure*}

We extend the analysis of Sec.~3.2 of the main paper to the remaining DiT-based SR backbones, applying the same zero-ablation procedure in which individual channels are set to zero and the resulting variation in reconstruction quality is measured across all metrics. Fig.~\ref{fig:zero_ablation_plot_overall} reports the full set of per-channel zero-ablation results for DiT4SR and TEASR, while Fig.~\ref{fig:qualitative_disruption} illustrates the corresponding visual effect of activation disruption on DiT4SR and TSD-SR. As it can be observed, both architectures reproduce the asymmetry described in the main paper: removing top-ranked channels degrades reconstruction quality sharply, whereas random and bottom-ranked removals leave it largely intact. These extended results are fully consistent with the main findings and indicate that the dominant role of high-activation channels is shared across all evaluated architectures rather than specific to a single model.

\section{Additional Analyses on Channel Selection}

\subsection{Importance of Channel Selection}

Table~\ref{tab:selection_ablation_all_models} compares different channel selection strategies across the three backbones. Selecting top-activation channels yields the best no-reference perceptual quality on all three models, and the only exception is TOPIQ~\cite{topiq} on DiT4SR, where bottom selection is marginally higher.

These results confirm that channels with the highest activation magnitude carry the most perceptually relevant information, and that modulating them leads to gains in perceptual quality. The picture on fidelity metrics is backbone-dependent: on TSD-SR and TEASR, intervening on less active channels primarily affects low-frequency content and preserves pixel-level fidelity, yielding the best LPIPS, whereas on DiT4SR top selection is the strongest choice on both fidelity metrics. This validates our choice of focusing on top-activation channels, as they provide the most effective control over perceptual reconstruction quality.

\begin{table}[t]
\centering
\footnotesize
\setlength{\tabcolsep}{2pt}

\resizebox{\linewidth}{!}{
\begin{tabular}{
@{}
l
cc
ccccc
@{}
}

\toprule

&
\multicolumn{2}{c}{\textbf{Fidelity}}
&
\multicolumn{5}{c}{\textbf{Perceptual Quality}}
\\

\cmidrule(lr){2-3}
\cmidrule(lr){4-8}

&
{\scriptsize SSIM $\uparrow$}
&
{\scriptsize LPIPS $\downarrow$}
&
{\scriptsize MANIQA $\uparrow$}
&
{\scriptsize MUSIQ $\uparrow$}
&
{\scriptsize CLIP-IQA $\uparrow$}
&
{\scriptsize TOPIQ $\uparrow$}
&
{\scriptsize LIQE $\uparrow$}
\\

\midrule

\rowcolor{lightgray}
\textbf{TSD-SR}~\cite{dong2025tsd}
& & & & & & & \\

Baseline
& 71.68 & 31.11 & 58.12 & 66.01 & 73.63 & 62.49 & 4.05 \\

Bottom
& 73.25 & 30.59 & 58.16 & 66.47 & 73.26 & 64.40 & 4.18 \\

Random
& 73.37 & \textbf{30.35} & 58.84 & 66.77 & 73.78 & 63.34 & 4.21 \\

\rowcolor{OurColor}
\textbf{Top (Ours)}
& \textbf{73.94} & 31.07 & \textbf{60.13} & \textbf{68.16}
& \textbf{76.32} & \textbf{67.39} & \textbf{4.47} \\

\midrule

\rowcolor{lightgray}
\textbf{DiT4SR}~\cite{duan2025dit4sr}
& & & & & & & \\

Baseline
& 61.02 & 43.72 & 60.88 & 65.20 & 69.36 & 58.50 & 4.08 \\

Bottom
& 64.10 & 38.82 & 62.78 & 65.64 & 71.34 & \textbf{59.27} & 4.10 \\

Random
& 62.60 & 39.58 & 61.65 & 65.03 & 71.21 & 58.97 & 4.01 \\

\rowcolor{OurColor}
\textbf{Top (Ours)}
& \textbf{66.26} & \textbf{37.18} & \textbf{62.99} & \textbf{66.06}
& \textbf{71.38} & 58.67 & \textbf{4.20} \\

\midrule

\rowcolor{lightgray}
\textbf{TEASR}~\cite{gao2026teasr}
& & & & & & & \\

Baseline
& 73.07 & 30.97 & 56.17 & 63.10 & 56.90 & 57.42 & 3.09 \\

Bottom
& 73.86 & 30.28 & 56.49 & 63.56 & 58.03 & 58.19 & 3.17 \\

Random
& 73.86 & \textbf{30.27} & 57.50 & 64.10 & 59.54 & 58.84 & 3.30 \\
\rowcolor{OurColor}
\textbf{Top (Ours)}
& \textbf{75.71} & 30.84 & \textbf{61.42} & \textbf{67.73}
& \textbf{65.39} & \textbf{65.14} & \textbf{4.10} \\

\bottomrule
\end{tabular}
}

\vspace{-0.2cm}

\caption{\textbf{Effect of the channel selection strategy on DRealSR across different backbones.}
All variants modulate $K=8$ channels per stream and block. Top-channel selection gives the
best no-reference perceptual quality on every backbone, the only exception being TOPIQ on
DiT4SR.}

\label{tab:selection_ablation_all_models}
\vspace{-0.6cm}

\end{table}

\subsection{Standard Deviation vs. Mean Absolute Activation}

In the main analysis (Sec.~3.2 of the main paper), we quantify channel importance using the mean absolute activation. While this choice captures the overall magnitude of activations, alternative statistics may provide different signals.

To this end, we consider a variant based on the standard deviation of activations across tokens. Specifically, for each channel $c$ of stream $s$ at block $\ell$, we define
\begin{equation}
    \tilde{a}_{i,s}^{(\ell)}(c) =
    \sqrt{\frac{1}{T_s} \sum_{t=1}^{T_s} \left(A_{i,s}^{(\ell)}(t,c) - \mu_{i,s}^{(\ell)}(c)\right)^2},
\end{equation}
where $\mu_{i,s}^{(\ell)}(c) = \frac{1}{T_s}\sum_{t=1}^{T_s} A_{i,s}^{(\ell)}(t,c)$ denotes the mean activation of channel $c$. We then rank channels according to $\tilde{a}_{i,s}^{(\ell)}(c)$ instead of $a_{i,s}^{(\ell)}(c)$, while keeping every subsequent step unchanged: the same Phase~1 and Phase~2 pipeline is applied, and channel ablations follow the same protocol.

Table~\ref{tab:std_vs_meanabs} reports the performance of the resulting trained models. Mean absolute activation yields slightly better overall performance than standard deviation, improving six of the seven metrics and trailing only on MANIQA, which supports its use as the channel-selection criterion.

\begin{table}[t]
\centering
\footnotesize
\setlength{\tabcolsep}{1.5pt}

\resizebox{\linewidth}{!}{
\begin{tabular}{
@{}
l
cc
ccccc
@{}
}

\toprule

&
\multicolumn{2}{c}{\textbf{Fidelity}}
&
\multicolumn{5}{c}{\textbf{Perceptual Quality}}
\\

\cmidrule(lr){2-3}
\cmidrule(lr){4-8}

&
{\scriptsize SSIM $\uparrow$}
&
{\scriptsize LPIPS $\downarrow$}
&
{\scriptsize MANIQA $\uparrow$}
&
{\scriptsize MUSIQ $\uparrow$}
&
{\scriptsize CLIP-IQA $\uparrow$}
&
{\scriptsize TOPIQ $\uparrow$}
&
{\scriptsize LIQE $\uparrow$}
\\

\midrule

Standard deviation
& 73.54
& 31.33
& \textbf{60.48}
& 68.02
& 76.31
& 66.69
& 4.44
\\

\rowcolor{OurColor}
\textbf{Mean absolute activation (Ours)}
& \textbf{73.94}
& \textbf{31.07}
& 60.13
& \textbf{68.16}
& \textbf{76.32}
& \textbf{67.39}
& \textbf{4.47}
\\

\bottomrule
\end{tabular}
}

\vspace{-0.2cm}

\caption{\textbf{Comparison between mean absolute activation and standard deviation as channel-selection criteria.}
All other components of the pipeline, including Phase~1 and Phase~2, are kept identical.
For each criterion, we select the optimal $K$ and report the performance of the resulting
trained model on DRealSR.}

\label{tab:std_vs_meanabs}
\vspace{-0.6cm}

\end{table}

\subsection{Effect of Modulation Parameterization}

Table~\ref{tab:scaleshift} compares different parameterizations of the modulation function.
Scale-only modulation already improves most metrics, whereas shift-only modulation provides
substantially smaller gains. Combining scale and shift yields the most consistent overall
performance, achieving the best result on MANIQA, MUSIQ, LIQE and CLIP-IQA, matching the
scale-only variant on LPIPS and trailing it by only $0.02$ SSIM~\cite{1284395}.
Removing the output constraints instead leads to an uneven behavior: although TOPIQ increases, SSIM
drops substantially and several other metrics deteriorate. A visual inspection makes the
failure mode explicit, as the unconstrained variant covers the whole image with a hallucinated
high-frequency weave-like texture that erases the actual content, which no-reference metrics
reward while every reference-based metric collapses. This indicates that unconstrained affine
parameters produce overly aggressive interventions, and that bounding both scale and shift
improves optimization stability and provides a more reliable cross-metric performance profile.

\begin{table}[t]
\centering
\footnotesize
\setlength{\tabcolsep}{1.pt}

\resizebox{\linewidth}{!}{
\begin{tabular}{
@{}
l
cc
ccccc
@{}
}

\toprule

&
\multicolumn{2}{c}{\textbf{Fidelity}}
&
\multicolumn{5}{c}{\textbf{Perceptual Quality}}
\\

\cmidrule(lr){2-3}
\cmidrule(lr){4-8}

&
{\scriptsize SSIM $\uparrow$}
&
{\scriptsize LPIPS $\downarrow$}
&
{\scriptsize MANIQA $\uparrow$}
&
{\scriptsize MUSIQ $\uparrow$}
&
{\scriptsize CLIP-IQA $\uparrow$}
&
{\scriptsize TOPIQ $\uparrow$}
&
{\scriptsize LIQE $\uparrow$}
\\

\midrule
\rowcolor{lightgray}
\textbf{TSD-SR}~\cite{dong2025tsd}
& 71.68
& 31.11
& 58.12
& 66.01
& 73.63
& 62.49
& 4.05
\\

\quad no constraint
& 38.33
& 60.54
& 54.22
& 62.10
& 72.68
& \textbf{70.24}
& 4.28
\\

\quad $+$ scale
& \textbf{73.96}
& \underline{31.07}
& \underline{60.04}
& \underline{68.09}
& \underline{75.98}
& 67.14
& \underline{4.44}
\\

\quad $+$ shift
& 72.05
& \textbf{30.93}
& 58.35
& 66.15
& 73.77
& 62.72
& 4.09
\\

\rowcolor{OurColor}
\quad\textbf{Scale\&Shift (Ours)}
& \underline{73.94}
& \underline{31.07}
& \textbf{60.13}
& \textbf{68.16}
& \textbf{76.32}
& \underline{67.39}
& \textbf{4.47}
\\

\bottomrule
\end{tabular}
}

\vspace{-0.2cm}

\caption{\textbf{Ablation on the modulation parameterization on DRealSR.}
\textbf{Bold} marks the best and \underline{underline} the second-best result per metric.
The bounded affine transformation used by \ours{} gives the best fidelity/perception
balance, being best or second best on every metric.}

\label{tab:scaleshift}
\vspace{-0.3cm}

\end{table}

\subsection{How Far Does the Effect Extend? A Sliding Window over Channel Ranks}
\label{sec:rank_window}

Comparing top-$K$ and bottom-$K$ selection characterizes the two ends of the
importance ranking, while random selection provides an unstructured control.
To examine the transition between these regimes, we slide a window of fixed
width $K=8$ over the Phase-1 ranking. For a starting rank $s$, we modulate the
channels ranked in $[s,s+7]$ and retrain the predictor from scratch using the
same architecture, objective, optimization schedule, and trainable parameter
count, so that the runs differ only in the identity of the eight selected
channels. The window $s=1$ corresponds to our main configuration.

\begin{table}[t]
\centering
\footnotesize
\setlength{\tabcolsep}{1.pt}

\resizebox{\linewidth}{!}{
\begin{tabular}{
@{}
lc
cc
ccccc
@{}
}

\toprule

&
&
\multicolumn{2}{c}{\textbf{Fidelity}}
&
\multicolumn{5}{c}{\textbf{Perceptual Quality}}
\\

\cmidrule(lr){3-4}
\cmidrule(lr){5-9}

{\scriptsize Channel ranks}
&
{\scriptsize mag}
&
{\scriptsize SSIM $\uparrow$}
&
{\scriptsize LPIPS $\downarrow$}
&
{\scriptsize MANIQA $\uparrow$}
&
{\scriptsize MUSIQ $\uparrow$}
&
{\scriptsize CLIP-IQA $\uparrow$}
&
{\scriptsize TOPIQ $\uparrow$}
&
{\scriptsize LIQE $\uparrow$}
\\

\midrule

Baseline (no mod.)
& $1.0\times$
& 71.68
& 31.11
& 58.12
& 66.01
& 73.63
& 62.49
& 4.05
\\

\midrule

Random-8
& --
& 73.37
& 30.35
& 58.84
& 66.77
& 73.78
& 63.34
& 4.21
\\

Bottom-8
& --
& 73.25
& 30.59
& 58.16
& 66.47
& 73.26
& 64.40
& 4.18
\\

\midrule

33-40
& $2.3\times$
& \textbf{74.83}
& \textbf{29.80}
& 58.81
& 66.85
& 73.47
& 63.29
& 4.28
\\

17-24
& $3.4\times$
& 73.98
& 30.55
& 59.90
& 67.60
& 74.57
& 65.31
& 4.38
\\

9-16
& $4.8\times$
& 73.95
& 30.66
& \textbf{60.36}
& 67.81
& 75.08
& 65.20
& 4.41
\\

3-10
& $9.6\times$
& 73.81
& 31.24
& 59.74
& 67.92
& 75.40
& 66.56
& 4.42
\\

2-9
& $13.8\times$
& 73.99
& 30.99
& 59.63
& 67.84
& 74.96
& 66.40
& 4.38
\\

\rowcolor{OurColor}
\textbf{\ours} (1-8)
& $23.0\times$
& 73.94
& 31.07
& 60.13
& \textbf{68.16}
& \textbf{76.32}
& \textbf{67.39}
& \textbf{4.47}
\\

\bottomrule
\end{tabular}
}

\vspace{-0.2cm}

\caption{\textbf{A sliding window over channel ranks.}
Each row modulates the eight channels at the specified positions in the Phase-1 importance
ranking, while keeping the predictor architecture, objective, optimization schedule, and
parameter budget fixed. \emph{mag} denotes the mean activation magnitude of the selected
channels relative to the per-block channel mean.}

\label{tab:rank_window}
\vspace{-0.3cm}

\end{table}

\begin{table*}[t]
\centering
\footnotesize
\setlength{\tabcolsep}{1.5pt}
\resizebox{\linewidth}{!}{
\begin{tabular}{
@{}
c                 
p{0.85em}         
l                 
p{1.2em}          
%
r l p{0.85em}     
r l p{1.2em}      
%
r l p{0.85em}     
r l p{0.85em}     
r l p{0.85em}     
r l p{0.85em}     
%
>{\columncolor{lightgray!50}}r   
>{\columncolor{lightgray!50}}l   
@{}
}

\toprule


&
&
&
&
\multicolumn{5}{c}{\textbf{Fidelity}}
&
&
\multicolumn{14}{c}{\textbf{Perceptual Quality}}
\\

\cmidrule(lr){5-9}
\cmidrule(lr){11-24}

&
&
&
&
\multicolumn{2}{c}{SSIM $\uparrow$}
&
&
\multicolumn{2}{c}{LPIPS $\downarrow$}
&
&
\multicolumn{2}{c}{MANIQA $\uparrow$}
&
&
\multicolumn{2}{c}{MUSIQ $\uparrow$}
&
&
\multicolumn{2}{c}{CLIP-IQA $\uparrow$}
&
&
\multicolumn{2}{c}{TOPIQ $\uparrow$}
&
&
\multicolumn{2}{c}{\cellcolor{lightgray!50}LIQE $\uparrow$}
\\

\midrule


&
&
TSD-SR~\cite{dong2025tsd}
&
&
68.84 & &
&
27.96 & &
&
63.03 & &
&
70.72 & &
&
72.55 & &
&
66.40 & &
&
4.19 &
\\

\rowcolor{OurColor}
\cellcolor{white}
&
\cellcolor{white}
&
\textbf{\quad$+$ \ours (Ours)}
&
&
\textbf{70.43} & \inc{1.59}
&
&
28.44 & \dec{0.48}
&
&
\textbf{64.64} & \inc{1.61}
&
&
\textbf{72.14} & \inc{1.42}
&
&
\textbf{75.35} & \inc{2.80}
&
&
\textbf{69.83} & \inc{3.43}
&
&
\cellcolor{lightgray!50}\textbf{4.69}
&
\cellcolor{lightgray!50}\inc{0.50}
\\

\cmidrule(lr){3-24}

&
&
DiT4SR~\cite{duan2025dit4sr}
&
&
66.05 & &
&
33.35 & &
&
59.56 & &
&
63.63 & &
&
62.24 & &
&
54.81 & &
&
3.63 &
\\

\rowcolor{OurColor}
\cellcolor{white}
&
\cellcolor{white}
&
\textbf{\quad$+$ \ours (Ours)}
&
&
\textbf{69.27} & \inc{3.22}
&
&
\textbf{28.51} & \inc{4.84}
&
&
\textbf{61.70} & \inc{2.14}
&
&
\textbf{64.78} & \inc{1.15}
&
&
\textbf{64.66} & \inc{2.42}
&
&
\textbf{55.19} & \inc{0.38}
&
&
\cellcolor{lightgray!50}\textbf{3.74}
&
\cellcolor{lightgray!50}\inc{0.11}
\\

\cmidrule(lr){3-24}

&
&
TEASR~\cite{gao2026teasr}
&
&
69.62 & &
&
27.28 & &
&
58.98 & &
&
67.31 & &
&
54.83 & &
&
60.29 & &
&
3.28 &
\\

\rowcolor{OurColor}
\cellcolor{white}
&
\cellcolor{white}
&
\textbf{\quad$+$ \ours (Ours)}
&
&
\textbf{71.59} & \inc{1.97}
&
&
28.61 & \dec{1.33}
&
&
\textbf{64.68} & \inc{5.70}
&
&
\textbf{71.05} & \inc{3.74}
&
&
\textbf{62.27} & \inc{7.44}
&
&
\textbf{68.08} & \inc{7.79}
&
&
\cellcolor{lightgray!50}\textbf{4.17}
&
\cellcolor{lightgray!50}\inc{0.89}
\\

\cmidrule(lr){3-24}

&
&
SinSR~\cite{wang2024sinsr}
&
&
73.85 & &
&
30.50 & &
&
54.34 & &
&
61.46 & &
&
62.86 & &
&
53.52 & &
&
3.22 &
\\

&
&
OSEDiff~\cite{wu2024one}
&
&
73.41 & &
&
29.20 & &
&
63.35 & &
&
69.09 & &
&
66.87 & &
&
62.48 & &
&
4.07 &
\\

&
&
PiSA-SR~\cite{sun2025pixel}
&
&
74.14 & &
&
26.72 & &
&
65.54 & &
&
70.16 & &
&
67.14 & &
&
63.75 & &
&
4.11 &
\\

\multirow{-11}{*}{%
    \rotatebox[origin=c]{90}{\textbf{RealSR}}%
}
&
&
Gram-SR~\cite{fdoronzio2026gramsr}
&
&
76.68 & &
&
23.05 & &
&
58.93 & &
&
62.47 & &
&
54.48 & &
&
52.48 & &
&
3.12 &
\\

\midrule


&
&
TSD-SR~\cite{dong2025tsd}
&
&
71.68 & &
&
31.11 & &
&
58.12 & &
&
66.01 & &
&
73.63 & &
&
62.49 & &
&
4.05 &
\\

\rowcolor{OurColor}
\cellcolor{white}
&
\cellcolor{white}
&
\textbf{\quad$+$ \ours (Ours)}
&
&
\textbf{73.94} & \inc{2.26}
&
&
\textbf{31.07} & \inc{0.04}
&
&
\textbf{60.13} & \inc{2.01}
&
&
\textbf{68.16} & \inc{2.15}
&
&
\textbf{76.32} & \inc{2.69}
&
&
\textbf{67.39} & \inc{4.90}
&
&
\cellcolor{lightgray!50}\textbf{4.47}
&
\cellcolor{lightgray!50}\inc{0.42}
\\

\cmidrule(lr){3-24}

&
&
DiT4SR~\cite{duan2025dit4sr}
&
&
61.02 & &
&
43.72 & &
&
60.88 & &
&
65.20 & &
&
69.36 & &
&
58.50 & &
&
4.08 &
\\

\rowcolor{OurColor}
\cellcolor{white}
&
\cellcolor{white}
&
\textbf{\quad$+$ \ours (Ours)}
&
&
\textbf{66.26} & \inc{5.24}
&
&
\textbf{37.18} & \inc{6.54}
&
&
\textbf{62.99} & \inc{2.11}
&
&
\textbf{66.06} & \inc{0.86}
&
&
\textbf{71.38} & \inc{2.02}
&
&
\textbf{58.67} & \inc{0.17}
&
&
\cellcolor{lightgray!50}\textbf{4.20}
&
\cellcolor{lightgray!50}\inc{0.12}
\\

\cmidrule(lr){3-24}

&
&
TEASR~\cite{gao2026teasr}
&
&
73.07 & &
&
30.97 & &
&
56.17 & &
&
63.10 & &
&
56.90 & &
&
57.42 & &
&
3.09 &
\\

\rowcolor{OurColor}
\cellcolor{white}
&
\cellcolor{white}
&
\textbf{\quad$+$ \ours (Ours)}
&
&
\textbf{75.71} & \inc{2.64}
&
&
\textbf{30.84} & \inc{0.13}
&
&
\textbf{61.42} & \inc{5.25}
&
&
\textbf{67.73} & \inc{4.63}
&
&
\textbf{65.39} & \inc{8.49}
&
&
\textbf{65.14} & \inc{7.72}
&
&
\cellcolor{lightgray!50}\textbf{4.10}
&
\cellcolor{lightgray!50}\inc{1.01}
\\

\cmidrule(lr){3-24}

&
&
SinSR~\cite{wang2024sinsr}
&
&
75.53 & &
&
35.02 & &
&
50.32 & &
&
57.16 & &
&
65.62 & &
&
52.99 & &
&
3.22 &
\\

&
&
OSEDiff~\cite{wu2024one}
&
&
78.35 & &
&
29.67 & &
&
58.97 & &
&
64.69 & &
&
69.60 & &
&
59.98 & &
&
3.94 &
\\

&
&
PiSA-SR~\cite{sun2025pixel}
&
&
77.97 & &
&
29.63 & &
&
61.64 & &
&
66.12 & &
&
69.84 & &
&
63.33 & &
&
4.06 &
\\

\multirow{-11}{*}{%
    \rotatebox[origin=c]{90}{\textbf{DRealSR}}%
}
&
&
Gram-SR~\cite{fdoronzio2026gramsr}
&
&
81.21 & &
&
26.03 & &
&
53.80 & &
&
56.66 & &
&
58.11 & &
&
50.83 & &
&
2.88 &
\\

\bottomrule
\end{tabular}
}
\vspace{-0.15cm}
\caption{
\textbf{Comparison with existing diffusion-based SR methods on RealSR and DRealSR.}
Each frozen DiT-based backbone is reported together with its \ours-adapted
counterpart, followed by four external competitors. For each metric, the score
of an adapted backbone is followed by its direction-normalized change with
respect to the corresponding frozen backbone
($\Delta+$: improvement; $\Delta-$: degradation).
Bold indicates an improvement over the corresponding backbone rather than the
best result in the column. LIQE is shaded in gray because it is explicitly
included in the training objective of \ours.
}

\label{tab:sota_two_datasets_vertical}
\vspace{-0.35cm}

\end{table*}

Table~\ref{tab:rank_window} reveals a gradual but metric-dependent transition
across the importance ranking, rather than a strictly monotonic decay. As the
window approaches the upper tail, MUSIQ, LIQE, CLIP-IQA, and TOPIQ generally
increase, reaching their highest values for ranks 1-8. MANIQA follows a
similar overall trend but peaks slightly earlier, at ranks 9-16. Thus, the
top-ranked window provides the strongest aggregate no-reference perceptual
performance, although it is not optimal for every individual metric.

At the other end of the sweep, ranks 33-40 still contain channels whose
activation magnitude is $2.3\times$ the per-block average. Their perceptual
gains are broadly comparable to those of random selection: relative to the
unmodulated baseline, they improve MUSIQ by $+0.84$ versus $+0.76$ for
Random-8, and TOPIQ by $+0.80$ versus $+0.85$. On CLIP-IQA, however, they
slightly degrade the baseline ($-0.16$) where Random-8 marginally improves it
($+0.15$). At the same time, ranks 33-40 achieve the
best SSIM and LPIPS in the entire sweep. These results show that lower-ranked
channels remain useful intervention points, but that they favor a different
performance profile. The extreme upper tail is therefore not the only effective region;
rather, it provides the greatest leverage for jointly improving a broad set of
no-reference perceptual-quality metrics, whereas lower-ranked windows tend to
favor fidelity-oriented metrics.

\section{Detailed Comparison with Existing Methods}

\tinytit{Positioning}
We compare our approach against recent state-of-the-art super-resolution models to contextualize the absolute performance level of our adapted backbones. Importantly, the methods considered here (SinSR~\cite{wang2024sinsr}, OSEDiff~\cite{wu2024one}, PiSA-SR~\cite{sun2025pixel}, and Gram-SR~\cite{fdoronzio2026gramsr}) do not exhibit massive activations in their architectures, and therefore cannot directly benefit from our modulation strategy. This comparison is thus intended to position our adapted models within the broader SR landscape, rather than to claim direct superiority of the modulation mechanism itself.

\tit{Results}
Results are reported in Table~\ref{tab:sota_two_datasets_vertical}, in which each frozen backbone is paired with its \ours-adapted counterpart and reported alongside the four external competitors. As it can be observed, our best-performing adapted model (TSD-SR $+$ \ours) achieves competitive or superior perceptual scores on both datasets, reaching $75.35$ CLIP-IQA and $4.69$ LIQE on RealSR and $76.32$ CLIP-IQA and $4.47$ LIQE on DRealSR, surpassing every compared method on these two reference-free metrics. Some competitors remain stronger elsewhere: Gram-SR~\cite{fdoronzio2026gramsr} achieves the best distortion-oriented scores on both datasets, and PiSA-SR~\cite{sun2025pixel} the best MANIQA on RealSR. Gram-SR sits at the opposite end of the perception-distortion spectrum, pairing the best SSIM and LPIPS with the lowest LIQE overall and the lowest CLIP-IQA among the external competitors. The gains are also unevenly distributed across our backbones: TEASR starts from the lowest CLIP-IQA and LIQE on both datasets and is the one that improves the most on them ($+8.49$ and $+1.01$ on DRealSR, $+7.44$ and $+0.89$ on RealSR), suggesting that the largest headroom lies where the frozen prior is perceptually weakest. Overall, our approach provides a better balance between perceptual quality and fidelity, in line with the perception-distortion trade-off~\cite{Blau_2018_CVPR}, confirming that selectively modulating dominant channels in DiT-based backbones is an effective strategy to reach competitive perceptual quality within the broader SR landscape.

\section{Additional Qualitative Results}

\begin{figure*}[!t]
    \centering
    \includegraphics[width=0.89\linewidth]{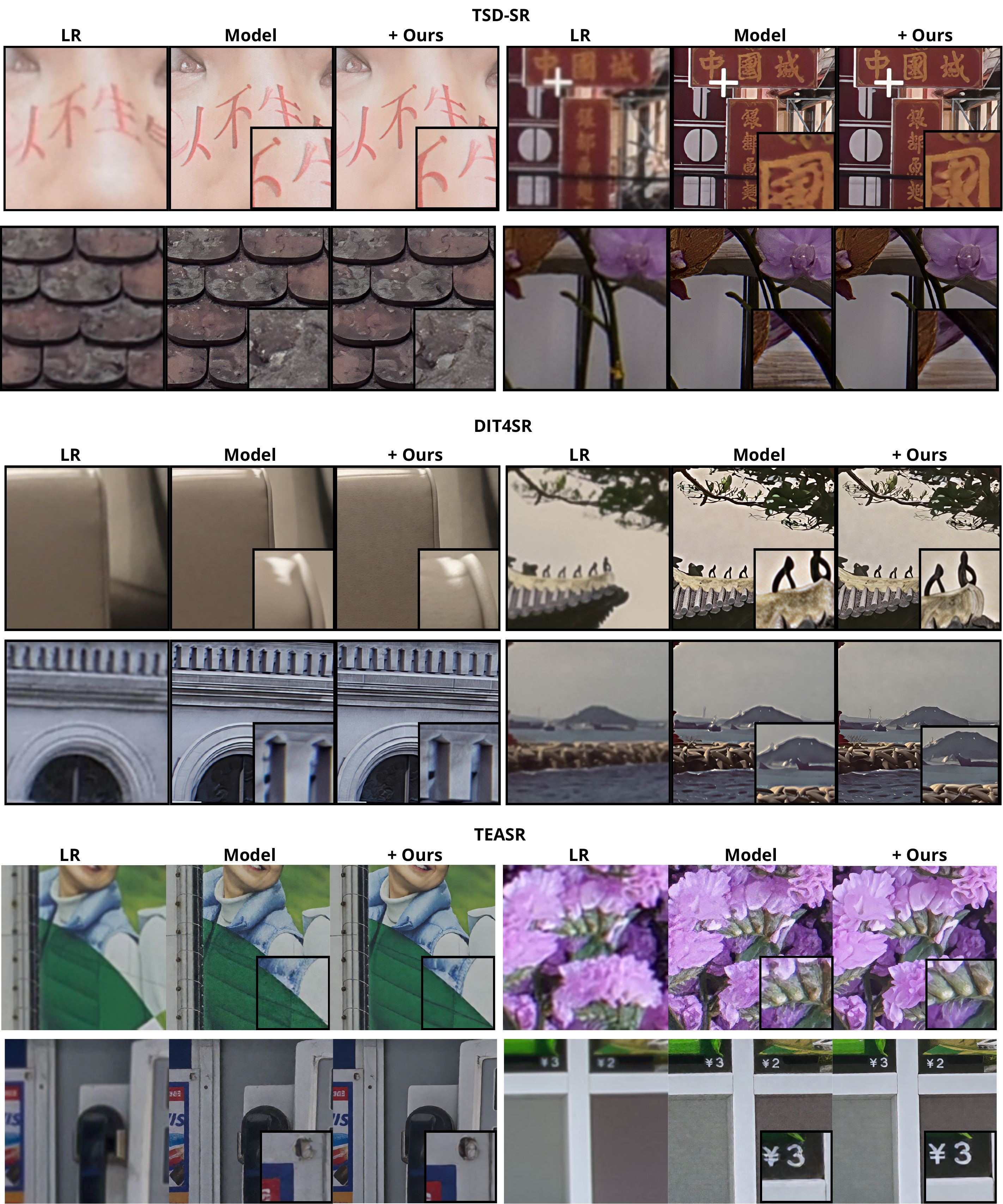}
    \vspace{-0.15cm}
    \caption{\textbf{Qualitative comparison across DiT-based SR backbones.} From left to right: low-resolution input, frozen baseline, and \ours. Results are shown for TSD-SR~\cite{dong2025tsd}, DiT4SR~\cite{duan2025dit4sr}, and TEASR~\cite{gao2026teasr} from top to bottom.}
    \label{fig:qualitatives_sup}
    \vspace{-0.3cm}
\end{figure*}

We provide additional qualitative comparisons in Fig.~\ref{fig:qualitatives_sup} to further illustrate the effect of the proposed modulation. Across different backbones and input images, \ours consistently produces sharper details and more perceptually plausible textures than the corresponding frozen baselines. Improvements are particularly visible in high-frequency regions, where the modulation enhances fine structures without introducing noticeable artifacts. At the same time, the global structure of the scene is preserved, indicating that the intervention remains localized and does not disrupt overall image consistency.

\section{Limitations}

\tinytit{Applicability of the selection procedure}
\ours relies on the presence of a sparse set of magnitude-dominant activation channels. Although this property is consistently observed across the three DiT-based SR backbones considered in our experiments, its strength may vary across architectures, and the proposed selection procedure may therefore be less effective when activation magnitude is distributed more uniformly across channels. Moreover, channel selection is backbone-specific, and Phase~1 must be repeated when changing the architecture or the checkpoint.

\tit{Scope and computational cost}
Our experiments focus on DiT-based super-resolution models and three evaluation datasets. Extending the approach to convolutional or U-Net-based backbones, to other restoration tasks, and to a broader range of real-world degradations remains an open direction. Finally, \ours reduces adaptation cost but does not accelerate the sampling procedure of the frozen backbone; consequently, inference remains dominated by the underlying SR model, especially for multi-step architectures such as DiT4SR~\cite{duan2025dit4sr}.

\end{document}